\documentclass[lettersize,journal]{IEEEtran}
\usepackage{amsmath,amsfonts}
\usepackage{algorithmic}
\usepackage{algorithm}
\usepackage{array}
\usepackage[caption=false,font=normalsize,labelfont=sf,textfont=sf]{subfig}
\usepackage{textcomp}
\usepackage{stfloats}
\usepackage{url}
\usepackage{verbatim}
\usepackage{graphicx}
\usepackage{cite}
\usepackage[table,xcdraw]{xcolor}
\usepackage{algorithm, bm}
\usepackage{algorithmic}
\usepackage{booktabs}
\usepackage{multirow}
\usepackage{makecell}

\begin{document}

\title{Query-Efficient Decision-based\\Black-Box Patch Attack}

\author{Zhaoyu Chen, Bo Li, Shuang Wu, Shouhong Ding, Wenqiang Zhang
\thanks{Corresponding authors are Bo Li and Wenqiang Zhang.}
\thanks{Zhaoyu Chen and Wenqiang Zhang are with Academy for Engineering and Technology, Fudan University, Shanghai, China, and also with Yiwu Research Institute of Fudan University, Yiwu, China.  The emails of these authors are: zhaoyuchen20@fudan.edu.cn, njumagiclibo@gmail.com, wqzhang@fudan.edu.cn.}}

\markboth{}%
{Shell \MakeLowercase{\textit{et al.}}: A Sample Article Using IEEEtran.cls for IEEE Journals}

\IEEEpubid{0000--0000/00\$00.00~\copyright~2021 IEEE}

\maketitle

\begin{abstract}
Deep neural networks (DNNs) have been showed to be highly vulnerable to imperceptible adversarial perturbations.
As a complementary type of adversary, patch attacks that introduce perceptible perturbations to the images have attracted the interest of researchers. 
Existing patch attacks rely on the architecture of the model or the probabilities of predictions and perform poorly in the decision-based setting, which can still construct a perturbation with the minimal information exposed -- the top-1 predicted label.
In this work, we first explore the decision-based patch attack.
To enhance the attack efficiency, we model the patches using paired key-points and use targeted images as the initialization of patches, and parameter optimizations are all performed on the integer domain.
Then, we propose a differential evolutionary algorithm named \textbf{DevoPatch} for query-efficient decision-based patch attacks.
Experiments demonstrate that DevoPatch outperforms the state-of-the-art black-box patch attacks in terms of patch area and attack success rate within a given query budget on image classification and face verification. 
Additionally, we conduct the vulnerability evaluation of ViT and MLP on image classification in the decision-based patch attack setting for the first time.
Using DevoPatch, we can evaluate the robustness of models to black-box patch attacks. We believe this method could inspire the design and deployment of robust vision models based on various DNN architectures in the future.
\end{abstract}


\section{Introduction}
Nowadays, deep neural networks (DNNs) have been employed as the fundamental techniques in the advancement of artificial intelligence in computer vision. Despite the success of DNNs, recent studies have identified that DNNs are vulnerable to adversarial examples \cite{L_BFGS}. By introducing maliciously crafted perturbations to the input images, these adversarial examples are able to evade and mislead DNNs. 
Consequently, studying the adversarial vulnerability of DNNs has emerged as an important research area, providing the opportunity to better understand and improve computer vision models.

\begin{figure}[t]
\centering
\begin{minipage}[b]{1\linewidth}
  \centering
  \centerline{\includegraphics[width=8.8cm]{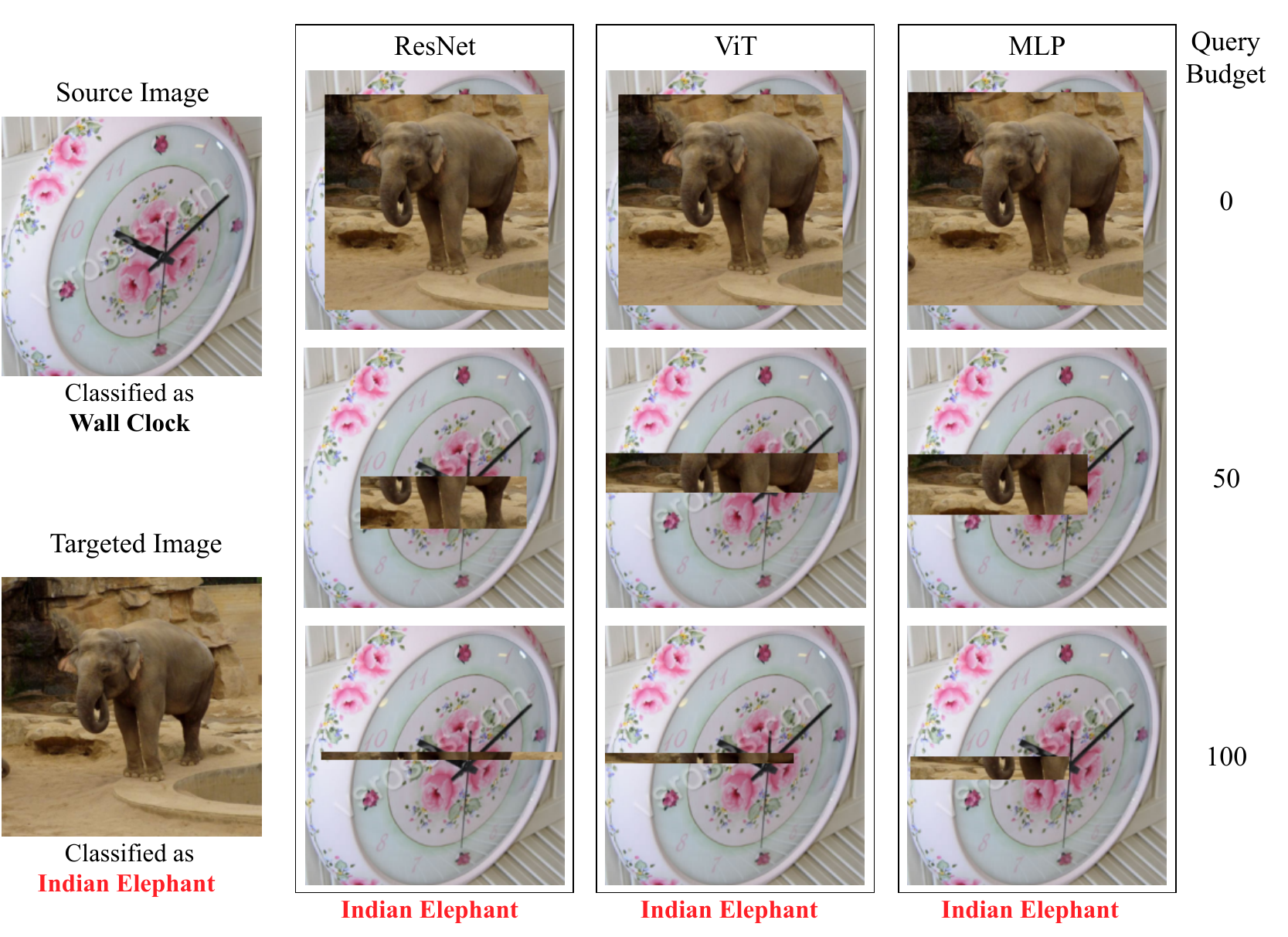}}
\end{minipage}
\caption{Introduction of DevoPatch. With regard to limited query budgets, we generate adversarial examples of patch attacks using DevoPatch applied to black-box models on image classification. As the number of queries increases, DevoPatch efficiently optimizes the quality of adversarial patches and achieves query-efficient decision-based patch attacks under a few query budgets.}
\label{fig:meaning}
\vspace{-0.4cm}
\end{figure}

Classical works \cite{L_BFGS, fgsm, ifgsm, pgd, cw, BP, autoattack} focus on studying the adversarial vulnerability of DNNs against virtually imperceptible perturbations that are constrained to have a small norm but are typically applied to the whole input image. Recently, as a complementary type of adversary, patch attacks that introduce perceptible (large norm) but localized perturbations to the images have attracted the interest of researchers.
Pioneering works\cite{adv_patch,lavan, ROA, LOAP,dapatch} perform patch attacks in the white-box setting: with full access to the model’s parameters and architectures, they can directly use gradient-based optimization to find successful adversarial examples. Due to the fact that most real-world applications do not publicly release the actual models they use, this attack scenario usually is less practical in real-world systems, e.g., attacking image analysis APIs \cite{DBLP:conf/icmla/HosseiniXP17} like Google Cloud Vision or self-driving cars\cite{DBLP:journals/corr/abs-1802-06430, PhysGAN, traffic_sign}.

\IEEEpubidadjcol

As a more practical scenario in real-world systems, black-box patch attacks have attracted a lot of attention in recent years. 
There are transfer-based attacks\cite{GenAP} and query-based attacks\cite{HPA, patchattack, AdvW, Sparse-RS} for black-box patch attacks, depending on whether the attacker needs to query the victim's machine learning model. 
Despite the fact that transfer-based attacks do not require query access to the model, it assumes the attacker has access to a large training set to create a carefully-designed substitute model\cite{hsja, QEBA, DFANet}, and there is no guarantee of success\cite{ensembledefense}. 
Query-based attacks assume that attackers can only query the target network and obtain its outputs (score or label) for a given input. 
According to the output information of the queried models, query-based attacks can be classified into two sub-categories: score-based setting which has access to the class probabilities of the model, and decision-based setting which solely relies on the top-1 predicted label.  
Significantly, decision-based settings present more practical threats to deployed systems and applications because an adversary is still capable of exploiting the very minimal information exposed -- the top-1 predicted label -- for constructing an adversarial perturbation.
Recently, some score-based patch attacks\cite{HPA, patchattack, AdvW, Sparse-RS} have been proposed. However, when these methods are applied to the decision-based setting, they hardly achieve high attack success rate and query efficiency because the information provided by labels is limited.

In this paper, we first explore the decision-based patch attack to better measure the practical threat of patch attacks. 
To successfully conduct decision-based black-box patch attacks, there are still non-negligible challenges to overcome:

\noindent\textbf{Complex Solution Space.} Performing patch attacks is extremely challenging since it involves searching for all possible positions, shapes, and perturbations of adversarial patches, which implies an enormous solution space. Moreover, unlike white-box scenarios or the score-based black-box setting, in the decision-based black-box setting, there is almost no valid information to guide the search direction.

\noindent\textbf{Query efficiency.} In the query-based setting, achieving high query efficiency with a high attack success rate is integral to adversarial objectives. Because: i) adversaries are able to carry out attacks at scale; ii) the cost of mounting the attack is reduced, and iii) adversaries are capable of bypassing defense systems that can recognize malicious activities as a fraud based on a pragmatically large number of successive queries with analogous inputs. Last but not least, the advantage of a smaller query budget is that it correlates to a lower cost of evaluation and research, which is useful for determining the robustness of the model to adversarial attacks.

To address the aforementioned issues, we propose a differential evolutionary algorithm named \textbf{DevoPatch} for query-efficient adversarial patch attacks in the decision-based black-box setting. 
Differential evolutionary algorithm is a black-box optimization algorithm that does not need to know the details of the model and is suitable for parameter search when information is limited.
Given the attack objective function, DevoPatch is able to optimize it in a black-box manner through queries only. 
To simplify the solution space, we restrict parameter optimization to the integer domain and carefully design a differential evolution algorithm based on the integer domain. Further, we model the patches using paired key-points and use targeted images as the initialization of patches.
Consequently, the query efficiency of DevoPatch is significantly improved. 
In addition, it is worth noting that some novel DNN architectures have recently emerged including the Vision Transformer (ViT) model \cite{vit} and Multi-layer Perceptron (MLP) based model \cite{mlp}. 
They demonstrate compelling performance, sometimes even outperforming classical convolutional architectures. Although a few studies have explored the vulnerability of ViT against imperceptible adversarial perturbations\cite{DBLP:journals/corr/abs-2103-14586,DBLP:journals/corr/abs-2103-15670}, the adversarial robustness of ViT and MLP under patch attacks has not been considered. This raises a critical security concern for the reliable deployment of real-world applications based on ViT and MLP models. Therefore, we extend our study scope and apply DevoPatch to ViT and MLP to better understand the vulnerability of a wide variety of DNNs under adversarial patch attacks. We illustrate an example patch attack with DevoPatch against ILSVRC2012 in Fig.~\ref{fig:meaning} on image classification. Extensive experiments on image classification and face verification demonstrate that DevoPatch is a query-efficient decision-based black-box patch attack.

We summarize our contributions and results below:
\begin{itemize}
    \item We first explore the decision-based patch attack, which can still construct a perturbation with the minimal information exposed -- the top-1 predicted label.
    \item To simplify the solution space, we model the patches using paired key-points and use targeted images as the initialization of patches, and parameter optimizations are all performed on the integer domain.
    \item We propose a novel patch attack -- DevoPatch -- an evolutionary algorithm capable of exploiting access to solely the top-1 predicted label from a model to search for an adversarial example,  whilst minimizing the image area that needs to be corrupted for a successful attack.
    \item Comprehensive experiments on image classification and face verification show that DevoPatch achieves considerably higher success rates compared to related work, while being more efficient in terms of the number of queries.
    \item We conduct the vulnerability evaluation of ViT and MLP on image classification in the decision-based black-box patch attack setting for the first time. We compare results with ResNet to assess the relative robustness of the ViT and MLP models.
\end{itemize}

The remainder of the paper is organized as follows. Section~\ref{sec:relatedwork} briefly reviews the literature related to adversarial examples and adversarial patches, white-box patch attacks, black-box patch attacks, and adversarial attacks with evolutionary algorithms. Section~\ref{sec:method} first introduces the definition of decision-based black-box patch attacks and then details the proposed
differential evolutionary patch attack. Section~\ref{sec:experiments} shows the experimental results to demonstrate the effectiveness of the proposed differential evolutionary patch attack. Firstly, we choose appropriate hyperparameters for DevoPatch. Afterward, we evaluate the adversarial robustness of several image classification and face recognition models. In Section~\ref{sec:discussion}, we further analyze the effects of adversarial patches on different DNN architectures. We summarize the paper in Section~\ref{sec:conclusions}.

\section{Related Work}
\label{sec:relatedwork}
In this section, we briefly review the literature related to adversarial examples and adversarial patches, white-box patch attacks, and black-box patch attacks. In the end, we also discuss adversarial attacks based on evolutionary algorithms.
\subsection{Adversarial Example and Adversarial Patch}  
The seminal works of Szegedy et al.\cite{L_BFGS} inspire an interest in studying adversarial vulnerability against imperceptible perturbations as a mean of understanding and improving deep neural networks. Since then, a majority of prior works\cite{fgsm, cw, ifgsm, pgd, BP, autoattack} have focused on attacking with small and imperceptible perturbations to the input, which can be regarded as \textit{the imperceptible adversarial attack}. Commonly these imperceptible perturbations are applied to the whole input image and are constrained by p-distances ($p \in \{0,2,\infty\}$) similarity measurement. Recently, as a complementary type of adversary, patch attacks that introduce perceptible (large norm) but localized perturbations to the images have emerged and attracted the interest of researchers. Patch attacks (or adversarial patches) can be regarded as \textit{the perceptible adversarial attack}.
The main aim of patch attacks is to minimize the perturbation within a continuous image region that needs to be corrupted to mislead a target machine learning model. Only a handful of works have investigated patch attacks and these works can be broadly categorized based on various degrees of adversarial access to a model. In this paper, we focus on black-box patch attacks because they are more practical and more threatening.

\subsection{White-box Patch Attack} 
In the white-box setting, an adversary has full knowledge and access to the model, including gradients and parameters.
GAP\cite{adv_patch} first creates universal, robust, targeted adversarial image patches in the real world and causes a classifier to output any target class in the white-box setting. Then LaVAN\cite{lavan} concentrates on investigating the blind spots of state-of-the-art image classifiers in the digital domain,  which crafts adversarial patches using an optimization-based approach with a modified loss function. Then\cite{ROA} and\cite{LOAP} introduce position search to improve the attack performance of adversarial patches. 
Due to the enormous solution space and the trade-off between computational cost and attack performance, adversarial patches are usually created with a fixed shape or location even under the white-box setting. 
Since then, adversarial patches have been used to attack self-driving cars\cite{DBLP:journals/corr/abs-1802-06430, PhysGAN, traffic_sign}, object detection\cite{rpattack, clork, Translucent_Patch} and face cognition\cite{adv_glass, adv_hat, adv_makeup}. 
However, white-box patch attacks are less practical, since most real-world applications do not release their models and cannot directly solve adversarial patches via gradients.
In this paper, we focus on black-box patch attacks because they are more threatening to real-world systems.

\subsection{Black-box Patch Attack} 
Black-box patch attacks can be either transfer-based\cite{GenAP} or query-based\cite{HPA, patchattack, AdvW, Sparse-RS}, depending on whether the attacker needs to query the victim's machine learning model. However, transfer-based attacks require access to large amounts of training data and require careful construction of surrogate models. It does not guarantee that the attack will be successful. In contrast, query-based attacks only require access to the output of the victimized model and have a higher attack success rate as the number of queries increases, which is more practical and more threatening. In the query-based attack, an adversary can access all or only one predicted score (score-based settings) or call out just the predicted labels (decision-based settings) for a given input. We need a query-efficient algorithm that helps reduce the cost of evaluating the robustness of DNNs since the attacker has to pay for each query.

Query-based patch attacks are first introduced in Hastings Patch Attack (HPA)\cite{HPA}. They do not optimize the pattern of the patches and instead use the monochrome patches. The position and shape of the rectangular patches are randomly searched using Metropolis-Hastings sampling.  
To improve the query efficiency of HPA, \cite{patchattack} first uses reinforcement learning to search the position and size of monochrome rectangular patches, called Monochrome Patch Attack (MPA). But monochrome patches usually lead to a very low attack success rate, especially for the targeted attack. They then use ImageNet training data to build a class-specific texture dictionary via style transfer\cite{styletransfer} to craft targeted patch attacks, termed Texture-based Patch Attack (TPA). However, in practical scenarios, it is impossible to obtain the whole training set data of the target black-box models. Adv-watermark (AdvW)\cite{AdvW} utilizes the basin hopping evolution algorithm to find a suitable position and transparency for the watermark to implement the patch attack. 
Patch-Rs\cite{Sparse-RS} proposes a random search framework and then designs an initialization scheme and a sampling distribution specific for adversarial patches. This outperforms previous works in both query efficiency and attack success rates.
Unfortunately, all of the aforementioned works are only designed for the score-based black-box setting. They perform poorly on the more challenging and restrictive decision-based attack (Experiment~\ref{sec:attack}). Further, decision-based settings present more practical threats to deployed systems.
To the best of our knowledge, we make the first investigation of the robustness of DNNs against patch attacks using a decision-based black-box setting.

\subsection{Adversarial Attacks with Evolutionary Algorithms} 
Notably, there are some related works\cite{onepixel, evoattack, sparseevo} which also leverage evolutionary algorithms to perform \textit{imperceptible attacks}.
These methods are all under the framework of evolutionary algorithms, with operations such as crossover and mutation. 
However, these methods cannot achieve query-efficient decision-based patch attacks due to the limitations of the modeling of adversarial examples.
One-pixel attack\cite{onepixel} generates one-pixel adversarial perturbations based on differential evolution. It is not effective when the number of perturbations is large because the number of queries to the model grows rapidly with respect to the number of perturbed pixels in patches. 
Evo-Attack\cite{evoattack} utilizes the covariance matrix adaptation evolution strategy to search the imperceptible perturbations but it cannot search for the location and shape of patches. 
SparseEvo\cite{sparseevo} models sparse perturbations as binary codes and solves them using genetic algorithms. However, this binary representation cannot define contiguous regions, thus making it impossible to model patches and perform patch attacks.
Our DevoPatch is based on the differential evolution algorithm, carefully designed in the integer domain to achieve query-efficient decision-based patch attacks. Consequently, we first construct a dimensionality-reduced solution space in which possible solutions (individuals of a population) are paired key-points in the integer domain. This is quite different from most current evolutionary attack methods.

\begin{figure*}[t]
\begin{minipage}[b]{1\linewidth}
  \centering
  \centerline{\includegraphics[width=16cm]{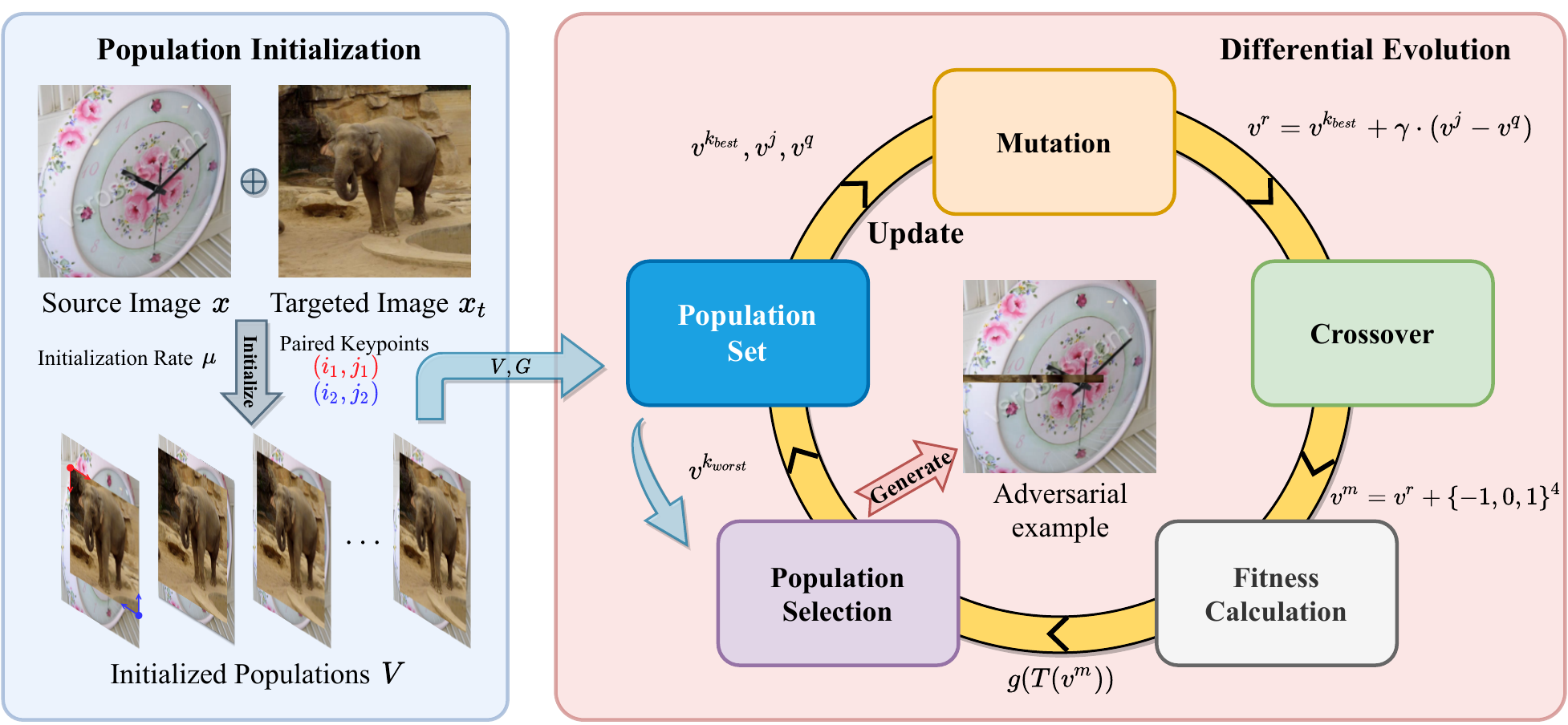}}
\end{minipage}
\caption{The pipeline of \textbf{DevoPatch}. The Population Initialization stage creates the initialized populations. During the differential evolution, by a combination of mutation, crossover, fitness calculation, and population selection, the population can improve over time to produce a satisfactory adversarial example.}
\label{fig:overview}
\end{figure*}

\section{Methodology}
\label{sec:method}
In this section, we first introduce the definition of decision-based black-box patch attacks and then detail the proposed differential evolutionary patch attack.

\subsection{Problem Definition}
Patch attacks are one of the most threatening types of adversarial examples that an adversary can arbitrarily modify the pixels of a continuous region, and the patch of this region leads machine learning models to make incorrect predictions. Here, we first give the formulation of patch attack on image classification. Face verification can be viewed as a binary classification task, similar to image classification. For an classifier $f:\bm{x} \rightarrow \bm{y}$, we are given a source image $\bm{x} \in \mathbb{R}^{C\times H \times W}$ and its corresponding ground truth label $\bm{y}$ from the label set $\mathbb{Y} = \{1, 2, ... ,K\}$ where $K$ denotes the number of classes. $C$, $W$, and $H$ denote the number of channels, height, and width of an image, respectively. In the setting of patch attacks, the adversarial patch is composed of adversarial perturbations $\delta \in \mathbb{R}^{C\times H \times W}$ and location masks $M\in \{0,1\}^{H \times W}$. Given a source image $\bm{x}$, we formulate the adversarial example $\bm{\Tilde{x}}$ as the combination of a source image $\bm{x}$, an adversarial patch $\delta$ and a location mask $M$:
\begin{equation}
    \bm{\Tilde{x}} = (I-M) \odot \bm{x} + M \odot \delta,
\end{equation}
where $\odot$ represents the element-wise Hadmard product and $I$ represents all-one matrices with the same dimension as $M$. 

In the decision-based black-box setting, our access is limited to its output label. For the targeted attack, the adversary perturbs the source image $\bm{x}$ so that the obtained adversarial example $\bm{\Tilde{x} }\in \mathbb{R}^{C\times H \times W}$ is misclassified as the desired class label $\bm{\Tilde{y}\in \mathbb{Y}}$. We refer to the desired class $\bm{\Tilde{y}}$ of the input $\bm{x}$ as the target class and its ground-truth class $\bm{y}$ as the source class. For the untargeted attack, the adversary perturbs the source image $\bm{x}$ to lead the output label of the classifier to any class label except the ground truth label $\bm{y}$, i.e. $\bm{\Tilde{y}} \in \mathbb{Y}$ where $\bm{\Tilde{y}} \neq \bm{y}$. In general, the patch attack (including targeted and untargeted settings) to find the best adversarial example $\bm{x^*}$ can be expressed as a constrained optimization problem: 
\begin{equation}
    \label{equ:objective}
    \bm{x^*} = \mathrm{arg} \min_{M,\delta} ||\bm{x}-\bm{\Tilde{x}}||_0 \quad s.t. \quad  f(\bm{x^*})=\bm{\Tilde{y}},
\end{equation}
where $||\cdot||_0$ denotes the number of perturbed pixels. For the patch not to be perceived, Eq.~\ref{equ:objective} aims to determine the perturbation and position with the constraint of a few perturbed pixels, which leads to a complex solution space and hampers the search. In addition, given the constraint and the fact that $f$ is not differentiable in the decision-based setting, the solution to the optimization problem is not trivial.

\subsection{Simplification on Solution Space} 
The enormous solution space on patch attacks is caused by all possible positions, shapes, and perturbations of the patch. 
A naive parametric search method can be directly used to solve this problem. Specifically, the parameter set $V$ is defined as a series of candidate solutions $\bm{v}$, represented by the coordinates and RGB values of each pixel. 
However, this naive application results in very inefficient queries\cite{onepixel}. 
Furthermore, in the decision-based black-box setting, there is little effective information to guide the search direction, thus further reducing the query efficiency.
To improve the query efficiency of decision-based attacks, we need to reduce the complexity of the solution space. 
Although in the decision-based setting, the black-box model can hardly provide effective information so the only information we can fully utilize is the target class $\bm{\Tilde{y}}$ of the targeted attack. 
To facilitate a parametric search method, instead of searching for parameters defining RGB values of each perturbed pixel, we consider that the perturbation $\delta$ can be replaced by a \textit{targeted image} $\bm{x_t}$. 
Targeted images are only required to be classified as target class by the black-box model and do not need to be i.i.d. with the training set (analyzed in Section~\ref{sec:source}).
Simultaneously, it is redundant to represent a patch with a coordinate set of perturbed pixels. For a patch, we only need to know a pair of points to formulate the location mask $M$ of the patch. Therefore, we vectorize each candidate solution in the parameter set $V$ as a 4-dimensional vector $\bm{v}=\{(i_1,j_1), (i_2,j_2)\}$ ($i_1 < i_2, j_1 < j_2$) where $i\in \mathbb{N}$ and $j\in \mathbb{N}$ denotes the coordinate of the paired key-points. Here, we employ a simple mapping function $T(\cdot)$ to re-formulate the location mask $M=T(v)$ and the adversarial example $\bm{\Tilde{x}}$:
\begin{equation}
    T(v)=\begin{cases}
        1,  &  \mathrm{if}\ 0 \leq i_1 < i_2 < H, 0 \leq j_1 < j_2 < W,\\
        0,  & \mathrm{otherwise},
    \end{cases}
\end{equation}
\begin{equation}
    \label{equ:generate}
    \bm{\Tilde{x}} = (I-M) \odot \bm{x} + M \odot \bm{x_t}.
\end{equation}

In general, we transform the original complex solution space into a coordinate programming problem for paired key-points on the integer domain. Interestingly, this strategy of simplification has been found to be extremely effective in a decision-based patch attack. Next, we need to design how to select and update suitable candidate solutions. 

\subsection{Differential Evolutionary Patch Attack}
\label{sec:DevoPatch}
In this section, we propose the \textbf{DevoPatch}, an efficient parametric search method based on the differential evolutionary algorithm that seeks a solution by iteratively improving upon potential solutions in search of a desirable one. 
In differential evolution, the population is the candidate solution and the population set is the parameter set.
We carefully design the differential evolution algorithm on the integer domain, including population initialization, mutation, and crossover.
Therefore, DevoPatch improves the differential evolution by simplifying solution spaces to the integer domain and the population can improve over time to produce a satisfactory result. 
Moreover, our search method employs the differential evolutionary algorithm without requiring any background knowledge of the target model, such as its architecture or parameters, to construct the fitness function. 
DevoPatch can be used to analyze and solve the non-trivial optimization problem in Eq.~\ref{equ:objective} in a black-box setting and can provide a possible remedy for complex solution space. 
The pipeline of DevoPatch is shown in Fig.~\ref{fig:overview} 

First, we give the definition of fitness calculation. Fitness is used to evaluate the quality of candidate solutions, mainly used in population initialization and population selection. In general, fitness function $g(\cdot)$ should reflect optimization objectives. In the score-based setting, since logits can be obtained, cross-entropy loss or margin loss\cite{cw} can be used to measure the quality of candidate solutions. In the decision-based setting, since only the predicted labels can be obtained, it is difficult for us to use the change of loss to measure the quality of new candidate solutions.
The loss only changes when the predicted label changes, which causes many potential candidates to be discarded.
Therefore, a fitness function is required to approximate the calculation of the loss function in the decision-based setting. Since our populations describe a paired key-point of the patch and the method uses targeted images as initialization, we consider an adversarial example with a smaller patch area would have better quality. Therefore, we formulate our fitness function as:
\begin{equation}
    \label{equ:fitness}
    g(\bm{\Tilde{x}})=\begin{cases}
        ||\bm{x-\Tilde{x}}||_0,  & \mathrm{if}\ f(\bm{\Tilde{x}})=\bm{\Tilde{y}}\\
        \infty,  & \mathrm{otherwise}.
    \end{cases}
\end{equation}
Although the fitness function is $l_0$ norm, other distance metrics are also feasible (further analyze in Section~\ref{sec:fitnessmeasure}).

Then, we initialize a population set of $p$ various candidate solutions named initialized population $\bm{v}^0$. In the population initialization, it is trivial to apply targeted images directly as initialization. The diversity among populations is conducive to improving query efficiency, so we introduce an \textit{initialization rate} $\mu$ to control the diversity of population initialization. Specifically, we first calculate height margin $\Delta h = \lfloor H\cdot \mu \rfloor $ and width margin $\Delta w = \lfloor W\cdot \mu \rfloor$ ($\Delta h \in \mathbb{N}, \Delta w \in \mathbb{N}$) as candidate domains. Then, every candidate solution is generated by only a randomly sample in the following condition:
\begin{equation}
    \begin{aligned}
        & i_1 \in [0,\ \Delta h), \ i_2 \in [H-\Delta h,\ H),\\& j_1 \in [0,\ \Delta w),\ j_2 \in [W-\Delta w,\ W].
    \end{aligned}
\end{equation}
Finally, if the fitness score of the initialized population $\bm{v}^0$ is not $\infty$ by using Eq.~\ref{equ:fitness}, the initialized population $\bm{v}^0$ will be successfully added to the population set $V$. Fitness scores are saved in a fitness score vector $G$ for each candidate solution. The population initialization is detailed in Algorithm~\ref{alg:Initialization}.

\begin{algorithm}[t]
\caption{Population Initialization Algorithm}
\label{alg:Initialization}
\textbf{Input}: source image $\bm{x}$, ground-truth label $\bm{y}$, targeted image $\bm{x_t}$, target label $\bm{\Tilde{y}}$, population  size $p$, initialization rate $\mu$ and model $f$\\
\textbf{Output}: $V, G$
\begin{algorithmic}[1] 
    \STATE $V \gets \emptyset,\ G \gets \infty$
      \FOR{$ i \gets 1,2,...,p$}
        \STATE $c \gets 0$
        \STATE $\Delta h \gets \lfloor H \cdot \mu \rfloor ,\ \Delta w \gets  \lfloor W \cdot \mu \rfloor$
        \WHILE{True}
            \STATE Generate $\bm{v^0}$ with $\Delta h, \Delta w$ using Eq. 6
            \STATE Generate $\bm{\Tilde{x}}$ with $\bm{v^0}, \bm{x_t}$ using Eq. 4
            \STATE Calculate $g(\bm{x^*})$ with $f(\bm{\Tilde{x}})$ using Eq. 5
            \IF{$f(\bm{\Tilde{x}}) = \bm{\Tilde{y}}$ \AND $g(\bm{x^*}) < G_i $  }
                \STATE $G_i \gets g(\bm{x^*})$
                \STATE $V \gets V \cup \{v^0\}$
                \STATE \textbf{break}
            \ENDIF
            \IF{$c > 10$}
                \STATE $\Delta h \gets 1,\ \Delta w \gets 1$
            \ENDIF
            \STATE $c \gets c + 1$
            
        \ENDWHILE
      \ENDFOR 
    \RETURN $V,G$
\end{algorithmic}
\end{algorithm}

Mutation is an important step in generating superior offspring (new candidate populations). Although the initialized population in the population initialization stage has a certain diversity, the overall difference is not significant, which will cause the next generation to be very similar and reduce query efficiency. In order to ensure the diversity of offspring, we introduce \textit{mutation rate} $\gamma$ to generate better offspring. Compared with the traditional differential evolutionary algorithm\cite{DE}, we need to ensure that the calculation is closed and the solution set of offsprings is an integer domain, so \textit{mutation rate} $\gamma$ must be an integer, i.e. $\gamma=1$. Specifically, when DevoPatch converges, the coordinate difference in candidate solutions is often only 1. If $\gamma$ is a real number (i.e. $\gamma=0.5$), it may be 0 after rounding to coordinates, resulting in no new offspring and reducing diversity. In practice, we first select the best $\bm{v}^{k_{best}}$ and two randomly selected candidate solutions $\bm{v}^j$, $\bm{v}^q$. Then, $\bm{v}^r$ is based on $\bm{v}^{k_{best}}$ plus $\gamma$ times the difference between $\bm{v}^j$ and $\bm{v}^q$. Formally, the mutation can be formulated as:
\begin{equation}
    \label{equ:mutation}
    \bm{v}^r = \bm{v}^{k_{best}} + \gamma \cdot (\bm{v}^j - \bm{v}^q).
\end{equation}

In order to increase the diversity of the generated population $\bm{v}^r$, crossover is introduced. The diversity of a population enables the exploration of the solution space for better individuals. Consequently, crossover operation is a crucial part of our method for further promoting population diversity, and every offspring after mutation can have the crossover. Unlike the traditional differential evolution algorithm~\cite{DE}, because the paired key-points of the modeling patches have an order relationship, it is impossible to directly perform crossover by element according to the probability. In practice, for any element in a candidate solution $\bm{v}^r$ after mutation, we randomly add a noise $\kappa = \{-1, 0, 1\}^4$ to it respectively, to help jump out of the local optimal solution. Therefore, crossover can be expressed as:
\begin{equation}
    \label{equ:crossover}
    \bm{v}^m = \bm{v}^{r} + \{-1, 0, 1\}^4.
\end{equation}

The evolution algorithm assumes that superior individuals are selected from a population and inferior individuals are eliminated. According to this assumption, individuals with better fitness scores are more likely to survive over time. Specifically, if an offspring has a smaller fitness score, it will also be better on Eq.~\ref{equ:objective} and be a better adversarial example. Hence, if a new offspring has a smaller fitness score than the worst offspring in the population, the worst offspring will be discarded and the new offspring will be selected in its place.

\begin{algorithm}[t]
\caption{DevoPatch}
\label{alg:DevoPatch}
\textbf{Input}: source image $\bm{x}$, ground-truth label $\bm{y}$, targeted image $\bm{x_t}$, target label $\bm{\Tilde{y}}$, query budget $N$, population  size $p$, initialization rate $\mu$, mutation rate $\gamma$, model $f$\\
\textbf{Output}: adversarial example $\bm{x^*}$ 
\begin{algorithmic}[1] 
    \STATE $V,G \gets \mathrm{PopulationInitialization} (f, \bm{x}, \bm{y}, \bm{x_t}, \bm{\Tilde{y}}, \mu, p)$
    \STATE $k_{best} \gets \mathrm{arg}\min_{k} (G),\ k_{worst} \gets \mathrm{arg}\max_{k} (G)$
      \FOR{$ i \gets 1,2,...,N$}
        \STATE Random sample $\bm{v^j, v^q}$ from $V \backslash \bm{v}^{k_{best}}$ 
        \STATE Initial $\bm{v^r}$ with $\bm{v^j, v^q}, \bm{v}^{k_{best}}$ using Eq.~\ref{equ:mutation}
        \STATE Generate $\bm{v^m}$ by random noises $\kappa$ using Eq.~\ref{equ:crossover}
        \STATE Generate $\bm{\Tilde{x}}$ with $\bm{v^m}, \bm{x_t}$ using Eq.~\ref{equ:generate}
        \STATE Calculate $g(\bm{x^*})$ with $f(\bm{\Tilde{x}})$ using Eq.~\ref{equ:fitness}
        \IF{$g(\bm{x^*}) < G_{k_{worst}}$ }
            \STATE $\bm{v^{k_{worst}}} \gets \bm{v^m}$
            \STATE $G_{k_{worst}} \gets g(\bm{x^*})$
        \ENDIF
        \STATE $k_{best} \gets \mathrm{arg}\min_{k} (G),\ k_{worst} \gets \mathrm{arg}\max_{k} (G)$
      \ENDFOR 
    \STATE Generate $\bm{x^*}$ with $\bm{v^{k_{best}}}, \bm{x_t}$ using Eq.~\ref{equ:generate}
    \RETURN $\bm{x^*}$
\end{algorithmic}
\end{algorithm}

Algorithm~\ref{alg:DevoPatch} summarizes the pipeline of DevoPatch. First, we obtain the initial population set and fitness scores through population initialization. Then, new offspring is generated by mutation and crossover to enhance diversity during each query. Next, the fitness score is calculated for the new population. Finally, the new population will be selected and updated according to the fitness score. Note that each time a new population is generated, we need perform boundary processing on the new population to ensure that $0 \leq i_1 < i_2 < H, 0 \leq j_1 < j_2 < W$.

\section{Experiments}
\label{sec:experiments}
In this section, we show the experimental results to demonstrate the effectiveness of the proposed differential evolutionary patch attack. First, we choose appropriate hyperparameters for DevoPatch. Then we evaluate the  adversarial robustness of several image classification models and face recognition models. We further conduct ablation studies and analyze the factors for the effectiveness of DevoPatch.

\subsection{Experimental Settings}
\subsubsection{Datasets} 
To evaluate the effectiveness of our method, we conduct experiments on image classification and face verification. For image classification, we follow\cite{patchattack} and conduct experiments on a challenging dataset, ILSVRC2012\cite{ilsvrc}, which has 1,000 object categories in total. For the evaluation sets, we randomly draw 1,000 correctly classified images from ILSVRC2012 validation set. 
Target images are also randomly chosen correctly classified images corresponding to target classes from the ILSVRC2012 validation set.
These selected images are evenly distributed among the 1,000 classes. For face verification, we select 400 pairs in dodging the attack, where each pair belongs to the same identity, and another 400 pairs in impersonation attack, where the images from the same pair are from different identities. The images are selected from LFW\cite{LFW} and CelebA\cite{CelebA}. Target images are also randomly chosen correctly recognized images corresponding to identities from LFW and CelebA. All the selected images can be correctly recognized by the face recognition models.

\subsubsection{Models} To evaluate the effectiveness of DevoPatch on different network architectures of image classification models, we select three different architecture models as threat models. For convolution-based models, we use a pre-trained ResNet-152 (ResNet)\cite{resnet}  for ILSVRC2012. For attention-based models, we select a pre-trained ViT-B-16/224 model (ViT-B)\cite{vit}. For multi-layer-based models, we select MLP-Mixer-B-16/224 model (Mixer-B)\cite{mlp}. We also study three face recognition models, including FaceNet\cite{FaceNet}, CosFace\cite{CosFace} and ArcFace\cite{ArcFace}, which all achieve over 99\% accuracies on the validation set. The threshold for the face recognition model is the one that achieves the highest accuracy on the validation set.

\subsubsection{Attack Methods} 
Decision-based settings present more practical threats to deployed systems because it is hard to get the score in the system.
To solve the practical issue of the score-based setting, our work first explores decision-based patch attacks, which can still construct a perturbation with the minimal information exposed -- the top-1 predicted label. Therefore, to reveal the issues of the score-based setting, all experimental comparisons are performed in the decision-based setting. For a fair comparison with score-based patch attacks, we choose HPA\cite{HPA}, MPA\cite{patchattack}, TPA\cite{patchattack}, Adv-watermark (AdvW)\cite{AdvW} and Patch-RS\cite{Sparse-RS} as the baseline under the decision-based setting. Inspired by \cite{DBLP:conf/icml/IlyasEAL18}, we leverage the label smoothing\cite{inceptionv2} to turn the hard-label into the score for the compared score-based methods without increasing the number of queries, where $\varepsilon=0.1$. Following\cite{Sparse-RS}, the patch areas of TPA, AdvW, and Patch-RS are fixed. For white-box patch attacks, we choose GAP\cite{adv_patch} as the baseline. GAP and DevoPatch share the same location mask $M$ and query (inference) budgets.

\subsubsection{Evaluation Metrics} 
Following\cite{patchattack} and \cite{Sparse-RS}, there are three metrics to measure the performance of black-box patch attacks. Patch area (\%) is the number of perturbed pixels divided by the total number of pixels of an image. To control how noticeable a patch is, we define an \textbf{Average Patch Area (APA)} as the average area across all successful attacks. To evaluate the efficiency of the patch attack, we calculate the \textbf{Average Number of Queries (ANQ)} over the images finished with patch attacks, followed by \cite{Sparse-RS}. Following decision-based adversarial attacks~\cite{QEBA,hsja}, we select the number of queries that reach the minimum value of the patch area during the query process as the calculated value of ANQ. Finally, a measure used to evaluate the adversarial robustness of a model is \textbf{Attack Success Rate (ASR)}. ASR (\%) is the ratio of adversarial examples that are successfully misrecognized.

\subsection{Effects of Hyperparameters}
\label{sec:ablation}
Here, we analyze the key factors of DevoPatch, including population size $p$, initialization rate $\mu$, and mutation rate $\gamma$ and fitness measure. All ablation experiments are performed on ResNet-152 for the image classification task.

\subsubsection{Population Size}
Table~\ref{tab:pop_size} shows the effect of different population sizes on performance, where $\mu=0.1$ and $\gamma=1$. As the population size $p$ gets larger, the ASR can still remain at 100\%, while the APA will decrease further and the ANQ will get greater. In particular, when $p=30$, its APA is 4.08\% less than $p=10$, but the ANQ is about 2 times larger. For the sake of query efficiency, we choose $p=10$.

\begin{table}[t]
\caption{Abalation study on population size $p$.}
\label{tab:pop_size}
\centering
\scalebox{0.95}{
\begin{tabular}{@{}c|ccc|ccc@{}}
\toprule
\multirow{2}{*}{Population Size $p$} & \multicolumn{3}{c|}{Untargeted Attack} & \multicolumn{3}{c}{Targeted Attack} \\ \cmidrule(l){2-7} 
 & ASR & APA & ANQ & ASR & APA & ANQ \\ \midrule
5 & 100.0 & 16.48 & 769.2 & 100.0 & 29.47 & 738.9 \\
10 & 100.0 & 12.16 & 1327.2 & 100.0 & 25.45 & 1317.2 \\
15 & 100.0 & 10.09 & 1918.6 & 100.0 & 23.89 & 1989.0 \\
20 & 100.0 & 9.36 & 2619.7 & 100.0 & 22.81 & 2678.4 \\
30 & 100.0 & 8.08 & 4163.6 & 100.0 & 22.29 & 4125.1 \\ \bottomrule
\end{tabular} 
}
\end{table}

\begin{table}[t]
\caption{Abalation study on initialization rate $\mu$.}
\label{tab:init_rate}
\centering
\scalebox{0.94}{
\begin{tabular}{@{}c|ccc|ccc@{}}
\toprule
\multirow{2}{*}{Initialization Rate $\mu$} & \multicolumn{3}{c|}{Untargeted Attack} & \multicolumn{3}{c}{Targeted Attack} \\ \cmidrule(l){2-7} 
 & ASR & APA & ANQ & ASR  & APA & ANQ \\ \midrule
0.05 & 100.0 & 12.69 & 1324.3 & 100.0 & 26.55 & 1322.5 \\
0.10 & 100.0 & 12.16 & 1327.2 & 100.0 & 25.45 & 1317.2 \\
0.15 & 100.0 & 11.64 & 1306.0 & 100.0 & 24.69 & 1287.7 \\
0.20 & 100.0 & 10.96 & 1320.5 & 100.0 & 24.63 & 1290.7 \\
0.25 & 100.0 & 10.19 & 1355.5 & 100.0 & 24.35 & 1263.1 \\
0.30 & 100.0 & 10.25 & 1391.1 & 100.0 & 23.94 & 1239.1 \\
\textbf{0.35} & \textbf{100.0} & \textbf{10.00} & \textbf{1349.7} & \textbf{100.0} & \textbf{23.78} & \textbf{1261.6} \\
0.40 & 100.0 & 9.99 & 1349.8 & 100.0 & 23.91 & 1260.4 \\ \bottomrule
\end{tabular}}
\end{table}

\begin{table}[t]
\centering
\caption{Abalation study on mutation rate $\gamma$.}
\label{tab:mutation_rate}
\centering
\scalebox{1.02}{
\begin{tabular}{@{}c|ccc|ccc@{}}
\toprule
\multirow{2}{*}{Mutation Rate $\gamma$} & \multicolumn{3}{c|}{Untargeted Attack} & \multicolumn{3}{c}{Targeted Attack} \\ \cmidrule(l){2-7} 
 & ASR & APA & ANQ & ASR  & APA & ANQ \\ \midrule
1 & 100.0 & 10.00 & 1349.7 & 100.0 & 23.78 & 1261.6 \\
2 & 100.0 & 7.75 & 6487.8 & 100.0 & 21.50 & 6205.2 \\
3 & 100.0 & 7.32 & 7294.9 & 100.0 & 21.24 & 7062.1 \\
4 & 100.0 & 6.91 & 7082.6 & 100.0 & 21.22 & 6909.2 \\ \bottomrule
\end{tabular}
}
\end{table}

\subsubsection{Initialization Rate}
Table~\ref{tab:init_rate} shows the effect of different initialization rates on performance, where $p=10$ and $\gamma=1$. As the initialization rate $\mu$ gradually increases, the APA will become less, and the ANQ will not change much. Due to the trade-off between areas and queries, we choose $\mu=0.35$. 

\subsubsection{Mutation Rate}
Table~\ref{tab:mutation_rate} shows the effect of different mutation rates on performance, where $\mu=0.35$ and $p=10$. Obviously, a larger mutation rate can indeed achieve a smaller adversarial patch, but it greatly increases the ANQ. Specifically, when $\gamma=4$, the APA is 3.09\% less than when $\gamma=1$, but the ANQ is 4 times greater. Considering query efficiency and average area, we choose $\gamma=1$.

\subsubsection{Convergence}
\begin{figure}[t]
\begin{minipage}[b]{0.49\linewidth}
  \centering
  \centerline{\includegraphics[width=4.5cm]{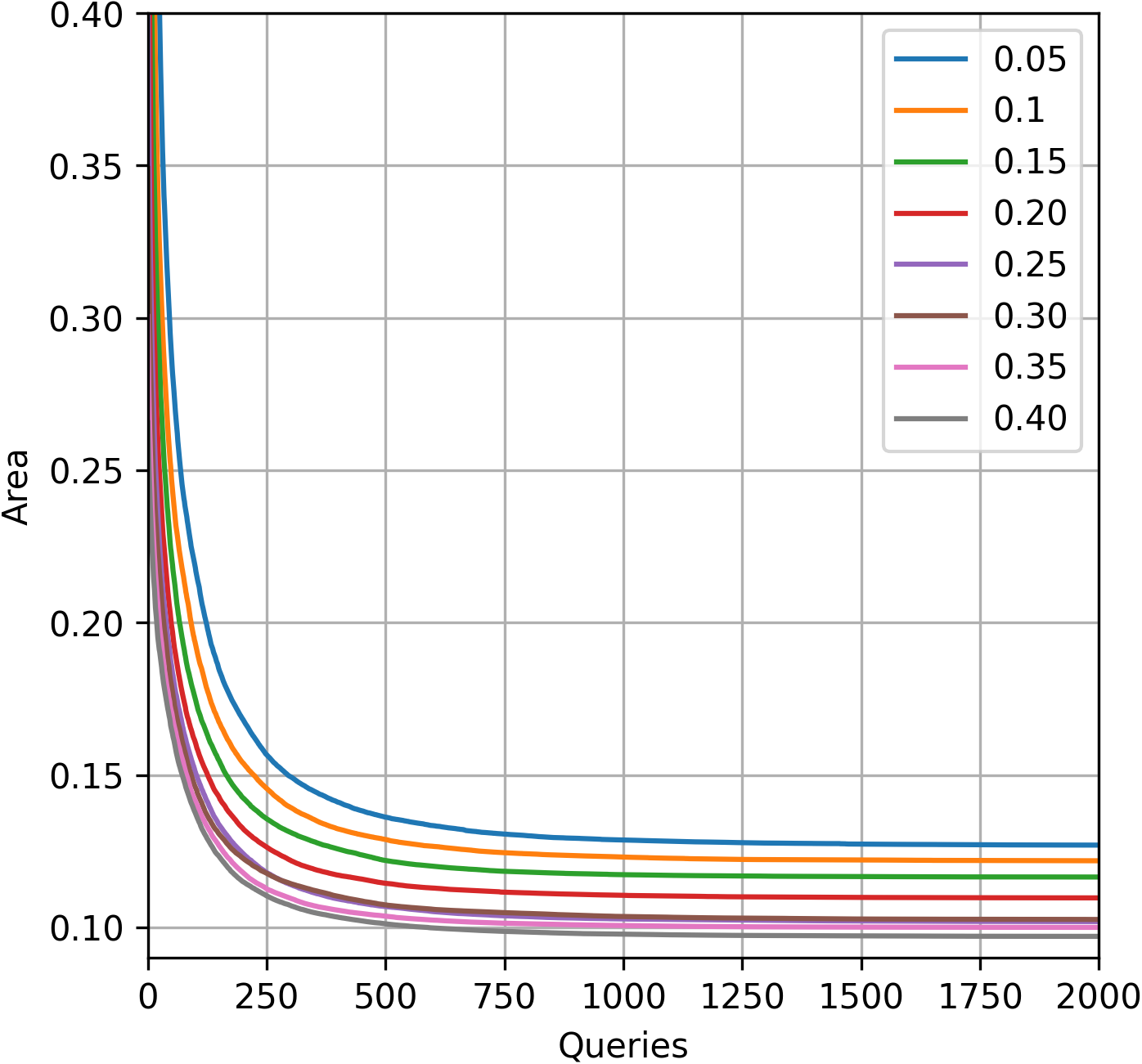}}
\centerline{\footnotesize (a) untargeted attack}
\end{minipage}
\hfill
\begin{minipage}[b]{0.49\linewidth}
  \centering
  \centerline{\includegraphics[width=4.5cm]{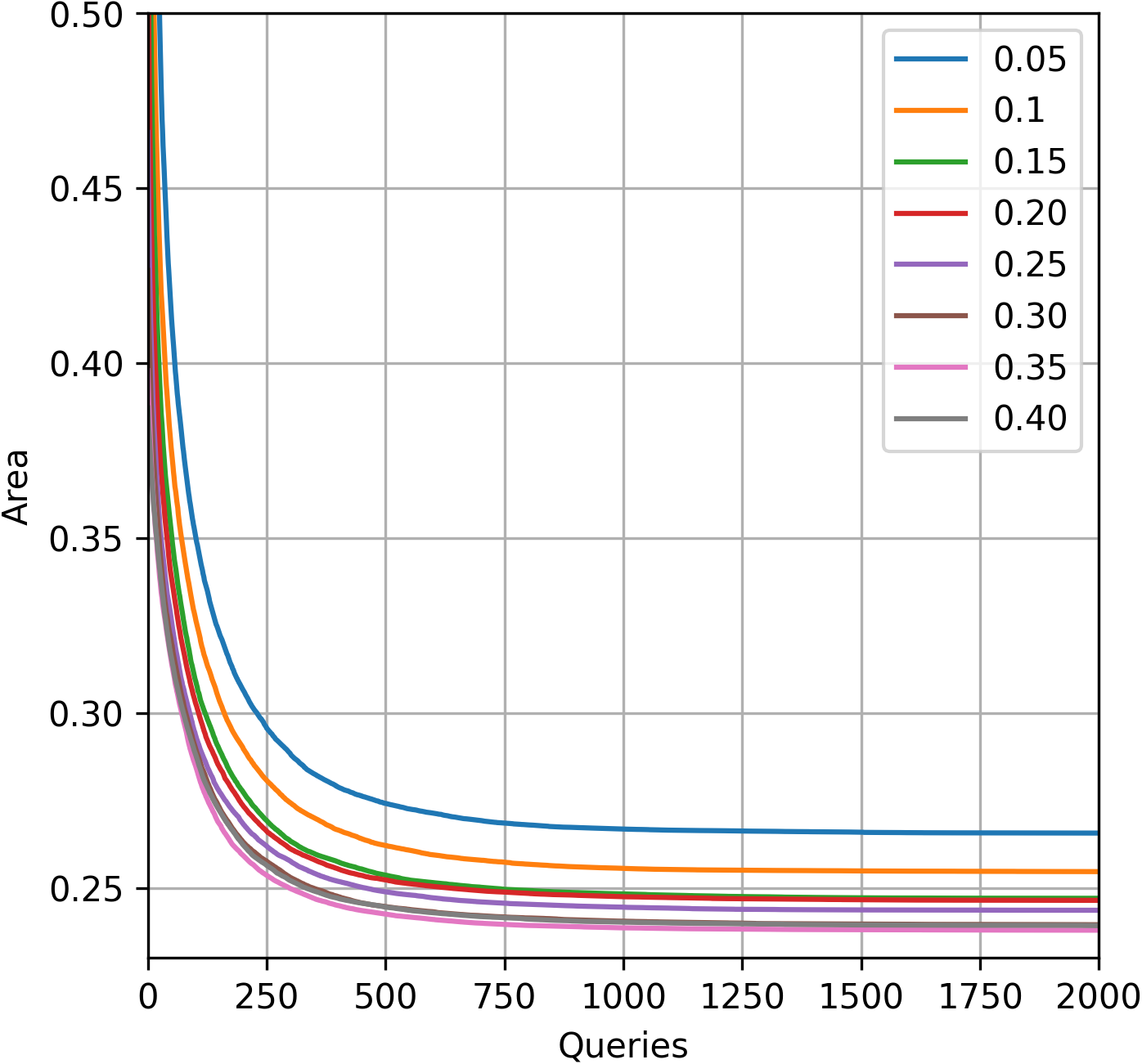}}
  \centerline{\footnotesize (b) targeted attack}
\end{minipage}
\caption{Convergence analysis. DevoPatch is query-efficient and can already generate high-quality adversarial examples when the ANQ is around 1,000.}
\label{fig:cover_mu}
\end{figure}

Further, we analyze the convergence of DevoPatch. Fig.~\ref{fig:cover_mu} describes the variation curve of area with query budget under different initialization rates $\mu$. Intuitively, our method converges quickly. When the number of queries is about 1,000, the best adversarial example has been solved.

\subsubsection{Fitness Measure} 
\label{sec:fitnessmeasure}
The fitness measure directly affects the efficiency of the solution.
In DevoPatch, we choose $l_0$ norm for the fitness calculation according to the optimization objective in Eq.~\ref{equ:objective}.
However, other norms can also be used to calculate Eq.~\ref{equ:fitness}. Table~\ref{tab:fitness} shows the ablation study about different norms on fitness calculation. $l_0$ norm consistently outperforms other norms in terms of queries and areas. The main reason may be that the optimization objectives of $l_0$ norm and Eq.~\ref{equ:objective} are consistent. Since other norms calculate the distance similarity with the image, they tend to place the patch in the place where the source image and the targeted image are similar, which will fall into the local optimal solution.

\begin{table}[t]
\centering
\caption{Abalation study on fitness measure.}
\label{tab:fitness}
\scalebox{0.93}{
\begin{tabular}{@{}c|c|ccc|ccc@{}}
\toprule
\multirow{2}{*}{Model} & \multirow{2}{*}{Norm} & \multicolumn{3}{c|}{Untargeted Attack} & \multicolumn{3}{c}{Targeted Attack} \\ \cmidrule(l){3-8} 
 &  & ASR  & APA  & ANQ & ASR  & APA & ANQ \\ \midrule
\multirow{3}{*}{ResNet} & 0 & 100.0 & \textbf{10.00} & \textbf{1349.7} & 100.0 & \textbf{23.78} & \textbf{1261.6} \\
 & 1 & 100.0 & 10.52 & 1395.3 & 100.0 & 24.64 & 1287.9 \\
 & 2 & 100.0 & 11.21 & 1394.7 & 100.0 & 25.77 & 1266.9 \\ \midrule
\multirow{3}{*}{ViT-B} & 0 & 100.0 & \textbf{13.84} & \textbf{1314.0} & 100.0 & \textbf{25.99} & \textbf{1256.6} \\
 & 1 & 100.0 & 14.79 & 1331.1 & 100.0 & 26.85 & 1285.4 \\
 & 2 & 100.0 & 15,40 & 1333.5 & 100.0 & 28.07 & 1290.5 \\ \midrule
\multirow{3}{*}{Mixer-B} & 0 & 100.0 & \textbf{14.76} & \textbf{1336.9} & 100.0 & \textbf{25.54} & \textbf{1274.6} \\
 & 1 & 100.0 & 15.29 & 1360.0 & 100.0 & 26.32 & 1286.0 \\
 & 2 & 100.0 & 16.39 & 1340.0 & 100.0 & 27.42 & 1282.1 \\ \bottomrule
\end{tabular}
}
\end{table}

\begin{table}[t]
\caption{Decison-based Black-box Patch Attacks on ILSVRC2012.}
\label{tab:black}
\centering
\scalebox{0.85}{
\begin{tabular}{@{}c|c|ccc|ccc@{}}
\toprule
\multirow{2}{*}{Model} & \multirow{2}{*}{Method} & \multicolumn{3}{c|}{Untargeted Attack} & \multicolumn{3}{c}{Targeted Attack} \\ \cmidrule(l){3-8} 
 &  & ASR  & APA  & ANQ & ASR & APA  & ANQ \\ \midrule
\multirow{6}{*}{ResNet} & HPA\cite{HPA} & 98.8 & 28.48 & 10000.0 & 0 & - & 50000.0 \\
 & MPA\cite{patchattack} & 99.5 & 14.10 & 10000.0 & 1.8 & 35.87 & 50000.0 \\
 & TPA\cite{patchattack} & 84.8 & 10.05 & 2768.0 & 82.0 & 24.12 & 12614.0 \\
 & AdvW\cite{AdvW} & 44.7 & 10.05 & 5913.0 & - & - & - \\
 & Patch-RS\cite{Sparse-RS} & 55.4 & 10.05 & 4754.4 & 0.1 & 24.12 & 49950.3 \\
 & Ours & \textbf{100.0} & \textbf{10.00} & \textbf{1349.7} & \textbf{100.0} & \textbf{23.78} & \textbf{1261.6} \\ \midrule
\multirow{5}{*}{ViT-B} & HPA\cite{HPA} & 98.1 & 34.04 & 10000.0 & 0 & - & 50000.0 \\
 & MPA\cite{patchattack} & 98.2 & 14.32 & 10000.0 & 2.1 & 66.38 & 49750.0 \\
 & TPA\cite{patchattack} & 82.3 & 14.06 & 3108.8 & 71.8 & 26.36 & 23794.6 \\
 & AdvW\cite{AdvW} & 30.6 & 14.06 & 7288.1 & - & - & - \\
 & Patch-RS\cite{Sparse-RS} & 42.1 & 14.06 & 6151.8 & 0.2 & 26.36 & 49940.2 \\
 & Ours & \textbf{100.0} & \textbf{13.84} & \textbf{1314.0} & \textbf{100.0} & \textbf{25.99} & \textbf{1256.6} \\ \midrule
\multirow{5}{*}{Mixer-B} & HPA\cite{HPA} & 87.6 & 39.53 & 10000.0 & 0 & - & 50000.0 \\
 & MPA\cite{patchattack} & 98.5 & 18.79 & 9850.0 & 2.2 & 80.58 & 49500.0 \\
 & TPA\cite{patchattack} & 96.1 & 15.08 & 1710.4 & 91.0 & 25.90 & 8037.0 \\
 & AdvW\cite{AdvW} & 40.7 & 15.08 & 6206.1 & - & - & - \\
 & Patch-RS\cite{Sparse-RS} & 63.5 & 15.08 & 3985.4 & 0.5 & 25.90 & 49801.2 \\
 & Ours & \textbf{100.0} & \textbf{14.76} & \textbf{1336.9} & \textbf{100.0} & \textbf{25.54} & \textbf{1274.6} \\ \bottomrule
\end{tabular}
}
\end{table}

\subsection{Attacks on Image Classification}
\label{sec:attack}
In this section, we compare the attack performance of various black-box patch attacks on image classification. We set the query budgets for untargeted and targeted attack to 10,000 and 50,000, respectively. The hyperparameters of DevoPatch are: $p=10$, $\mu=0.35$, $\gamma=1$. For targeted attacks, we consider a randomly chosen correctly classified image corresponding to the target class $\bm{\Tilde{y}}$ from the dataset. For untageted attacks, we use a randomly chosen correctly classified image corresponding to the random class except the ground-truth label from the dataset, followed by \cite{patchattack}. 
DevoPatch takes about 6.33 hours to perform 10,000 queries on 1,000 images on ResNet-152, based on an NVIDIA Tesla V100.
The experimental results against decision-based patch attacks in untargeted and targeted setting on ILSVRC2012 are summarized in Table~\ref{tab:black}. The experimental results show that our DevoPatch consistently outperforms HPA, MPA, TPA, AdvW and Patch-RS in terms of queries and patch areas with a higher ASR, which shows the effectiveness of DevoPatch. 

\begin{figure*}[t]
\begin{minipage}[b]{1\linewidth}
  \centering
  \centerline{\includegraphics[width=18cm]{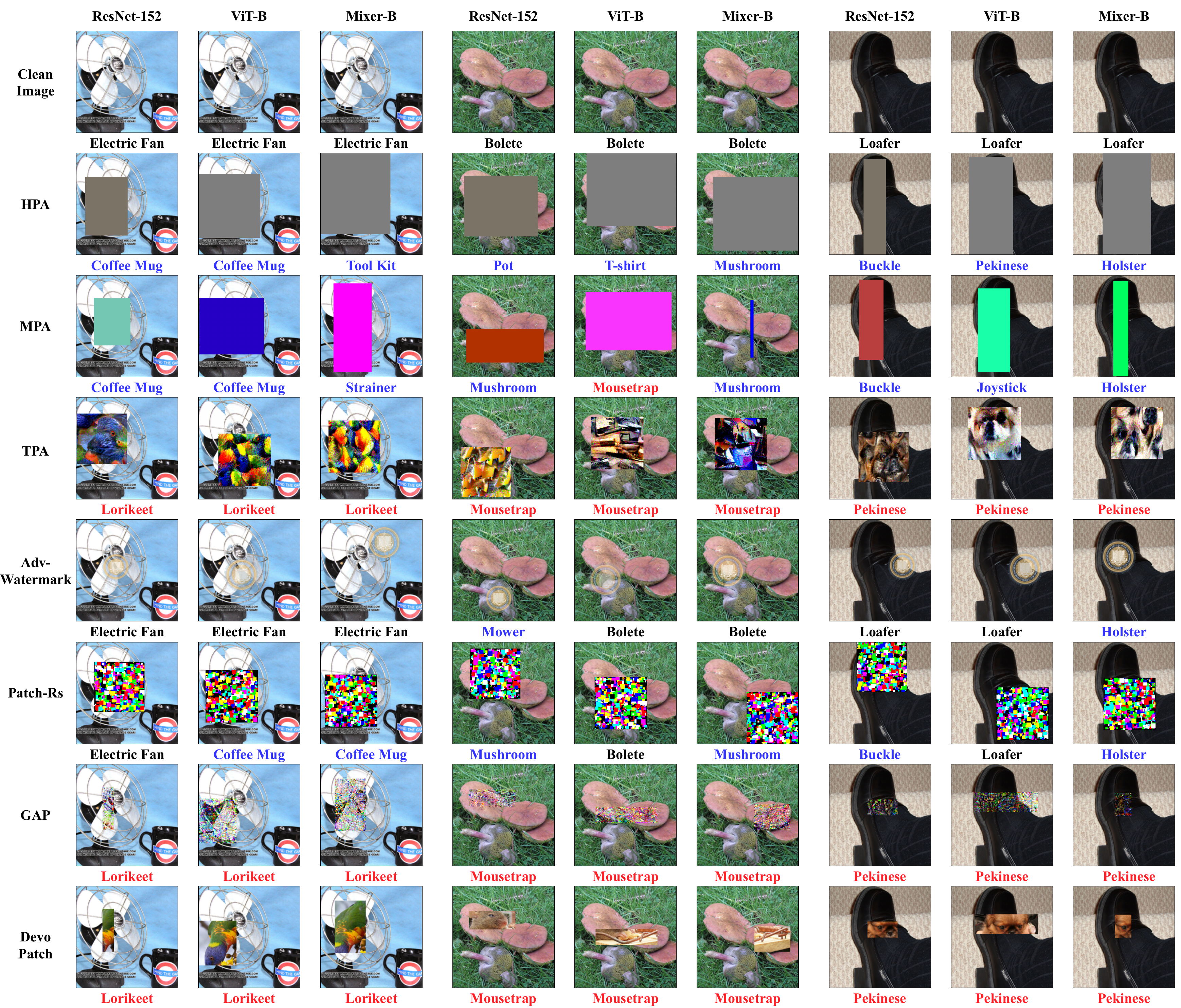}}
\end{minipage}
\caption{Visualization of patch attacks in the targeted setting on different network architectures. The labels below the image represent the predicted classes. Black, red and blue labels represent ground-truth labels, target classes, and the classes after the targeted attack has failed, respectively. DevoPatch successfully achieves the targeted attack of all examples with a small patch area.}
\label{fig:vis}
\end{figure*}

In the untargeted setting, TPA, AdvW, and Patch-RS achieve the trade-off on ASR and ANQ. Although the label returns little information, TPA achieves higher ASR in the decision-based setting due to its strong texture prior. Although HPA has a high ASR, its ANQ and APA are extremely large. MPA achieves sub-optimal ASR, but it is inefficient and always uses the whole query budget since it takes 10,000 queries and chooses the best one. DevoPatch achieves 100\% ASR with one-seventh of MPA on ANQ under a smaller average area. 
Under the more challenging targeted setting, due to the lack of effective information about the target class, HPA, MPA, AdvW, and Patch-RS are almost useless. Because TPA has a texture prior, it still has a high ASR, but the ANQ is extremely high. Because of our simplification of solution space, our DevoPatch outperforms TPA by 18.0\%, 28.2\%, and 9.0\% ASR on ResNet, ViT, and MLP, respectively, while the average queries are only one-tenth, one-twentieth and one-seventh of TPA. 
Also, we choose to compare with GAP, the most basic white-box patch attack. We expect DevoPatch to reach the lower bound of white-box patch attacks in attack performance. Table~\ref{tab:transfer} illustrates that DevoPatch achieves ASR equivalent to GAP. Both black-box and white-box experiments show that DevoPatch is a query-efficient decision-based black-box patch attack with high attack performance against different network architectures on image classification.

We provide the visualization of patch attacks on different network architectures as shown in Fig.~\ref{fig:vis}. The labels below the image indicate the predicted classes. Labels in black, red, and blue represent the ground truth, target classes, and the classes after the targeted attack has failed, respectively. HPA and MPA use gray or colored patches to achieve attacks (the second and third rows), but their patch area is large and it is difficult to achieve targeted attacks. TPA uses the ImageNet pre-trained texture dictionary for patch attack (shown in the fourth row) and has a higher ASR in the decision-based setting, but its area is also larger. AdvW selects pre-defined logos for patch attacks, but it is difficult to implement targeted attacks because logos have little category information (shown in the fifth row). Patch-Rs is based on a random search framework (shown in the sixth row), but it is difficult to implement targeted attacks because the top-1 labels have too little information. Our DevoPatch achieves the query-efficient attack in a decision-based setting with a smaller area and higher ASR.

\begin{table}[t]
\centering
\caption{Comparisons with white-box patch attacks on ASR (\%).}
\label{tab:transfer}
\scalebox{0.81}{
\begin{tabular}{@{}c|c|ccc|ccc@{}}
\toprule
\multirow{2}{*}{Source Model} & \multirow{2}{*}{Method} & \multicolumn{3}{c|}{Untargeted Attack} & \multicolumn{3}{c}{Targeted Attack} \\ \cmidrule(l){3-8} 
 &  & ResNet & ViT-B & Mixer-B & ResNet & ViT-B & Mixer-B \\ \midrule
\multirow{2}{*}{ResNet} & GAP\cite{adv_patch} & 100 & 5.3 & 9.3 & 100 & 0 & 0.2 \\
 & Ours & 100 & \textbf{7.5} & \textbf{10.3} & 100 & \textbf{10.4} & \textbf{14.6} \\ \midrule
\multirow{2}{*}{ViT-B} & GAP\cite{adv_patch} & 19.6 & 100 & 15.5 & 0 & 100 & 0 \\
 & Ours & \textbf{24.7} & 100 & \textbf{15.8} & \textbf{18.2} & 100 & \textbf{18.2} \\ \midrule
\multirow{2}{*}{Mixer-B} & GAP\cite{adv_patch} & 21.6 & 11.0 & 100 & 0 & 0 & 100 \\
 & Ours & \textbf{30.5} & \textbf{15.7} & 100 & \textbf{20.3} & \textbf{15.5} & 100 \\ \bottomrule
\end{tabular}
}
\end{table}

\begin{table*}[t]
\centering
\caption{Decison-based Black-box Patch Attacks on face verification.}
\label{tab:face}
\scalebox{1}{
\begin{tabular}{@{}cc|cccccc|cccccc@{}}
\toprule
\multicolumn{2}{c|}{Dataset} & \multicolumn{6}{c|}{LFW} & \multicolumn{6}{c}{CelebA} \\ \midrule
\multicolumn{1}{c|}{\multirow{2}{*}{Model}} & \multirow{2}{*}{Method} & \multicolumn{3}{c|}{Dodging Attack} & \multicolumn{3}{c|}{Impersonation Attack} & \multicolumn{3}{c|}{Dodging Attack} & \multicolumn{3}{c}{Impersonation Attack} \\ \cmidrule(l){3-14} 
\multicolumn{1}{c|}{} &  & ASR & APA & \multicolumn{1}{c|}{ANQ} & ASR & APA & ANQ & ASR & APA & \multicolumn{1}{c|}{ANQ} & ASR & APA & ANQ \\ \midrule
\multicolumn{1}{c|}{\multirow{6}{*}{ArcFace}} & HPA\cite{HPA} & 100 & 25.81 & \multicolumn{1}{c|}{10000.0} & 4.25 & 22.47 & 50000.0 & 100 & 10.78 & \multicolumn{1}{c|}{10000.0} & 32.75 & 15.28 & 50000.0 \\
\multicolumn{1}{c|}{} & MPA\cite{patchattack} & 100 & 14.29 & \multicolumn{1}{c|}{10000.0} & 5.00 & 17.24 & 50000.0 & 100 & 4.80 & \multicolumn{1}{c|}{10000.0} & 38.00 & 8.40 & 50000.0 \\
\multicolumn{1}{c|}{} & TPA\cite{patchattack} & 80.75 & 11.51 & \multicolumn{1}{c|}{3530.0} & 52.5 & 12.76 & 27352.5 & 86.50 & 4.98 & \multicolumn{1}{c|}{2754.0} & 58.50 & 7.66 & 25030.0 \\
\multicolumn{1}{c|}{} & AdvW\cite{AdvW} & 5.75 & \multicolumn{1}{c}{11.51} & \multicolumn{1}{c|}{9508.3} & - & - & - & 5.00 & \multicolumn{1}{c}{4.98} & \multicolumn{1}{c|}{9549.8} & - & - & - \\
\multicolumn{1}{c|}{} & Patch-Rs\cite{Sparse-RS} & 24.00 & 11.51 & \multicolumn{1}{c|}{7837.6} & 1.00 & 12.76 & 49513.9 & 56.00 & 4.98 & \multicolumn{1}{c|}{4838.3} & 16.00 & 7.66 & 42597.1 \\
\multicolumn{1}{c|}{} & Ours & \textbf{100} & \textbf{11.13} & \multicolumn{1}{c|}{\textbf{1016.4}} & \textbf{100} & \textbf{12.28} & \textbf{960.0} & \textbf{100} & \textbf{4.71} & \multicolumn{1}{c|}{\textbf{1043.4}} & \textbf{100} & \textbf{7.48} & \textbf{992.9} \\ \midrule
\multicolumn{1}{c|}{\multirow{6}{*}{CosFace}} & HPA\cite{HPA} & 100 & 14.14 & \multicolumn{1}{c|}{10000.0} & 10.75 & 15.74 & 50000.0 & 100 & 3.27 & \multicolumn{1}{c|}{10000.0} & 43.75 & 4.73 & 50000.0 \\
\multicolumn{1}{c|}{} & MPA\cite{patchattack} & 100 & 8.36 & \multicolumn{1}{c|}{10000.0} & 9.00 & 14.57 & 50000.0 & 100 & 2.71 & \multicolumn{1}{c|}{10000.0} & 59.50 & 4.82 & 50000.0 \\
\multicolumn{1}{c|}{} & TPA\cite{patchattack} & 76.5 & 7.29 & \multicolumn{1}{c|}{3741.0} & 56.6 & 12.05 & 25827.5 & 88.50 & 2.69 & \multicolumn{1}{c|}{2428.0} & 67.00 & 4.92 & 20177.5 \\
\multicolumn{1}{c|}{} & AdvW\cite{AdvW} & \multicolumn{1}{c}{8.00} & \multicolumn{1}{c}{7.29} & \multicolumn{1}{c|}{9303.0} & - & - & - & 13.00 & \multicolumn{1}{c}{2.69} & \multicolumn{1}{c|}{8742.7} & - & - & - \\
\multicolumn{1}{c|}{} & Patch-Rs\cite{Sparse-RS} & 55.25 & 7.29 & \multicolumn{1}{c|}{5024.3} & 5.5 & 12.05 & 47744.4 & 80.00 & 2.69 & \multicolumn{1}{c|}{2504.1} & 38.75 & 4.92 & 31729.5 \\
\multicolumn{1}{c|}{} & Ours & \textbf{100} & \textbf{7.02} & \multicolumn{1}{c|}{\textbf{1040.7}} & \textbf{100} & \textbf{11.40} & \textbf{1025.2} & \textbf{100} & \textbf{2.56} & \multicolumn{1}{c|}{\textbf{1041.7}} & \textbf{100} & \textbf{4.63} & \textbf{1080.2} \\ \midrule
\multicolumn{1}{c|}{\multirow{6}{*}{FaceNet}} & HPA\cite{HPA} & 100 & 15.13 & \multicolumn{1}{c|}{10000.0} & 16.00 & 18.05 & 50000.0 & 100 & 8.68 & \multicolumn{1}{c|}{10000.0} & 23.75 & 17.27 & 50000.0 \\
\multicolumn{1}{c|}{} & MPA\cite{patchattack} & 100 & 10.46 & \multicolumn{1}{c|}{10000.0} & 18.50 & 18.97 & 50000.0 & 100 & 5.50 & \multicolumn{1}{c|}{10000.0} & 31.50 & 13.37 & 50000.0 \\
\multicolumn{1}{c|}{} & TPA\cite{patchattack} & 87.50 & 6.56 & \multicolumn{1}{c|}{2748.0} & 76.00 & 12.69 & 16027.5 & 78.25 & 3.52 & \multicolumn{1}{c|}{3889.0} & 71.50 & 7.22 & 19640.0 \\
\multicolumn{1}{c|}{} & AdvW\cite{AdvW} & 18.75 & \multicolumn{1}{c}{6.56} & \multicolumn{1}{c|}{8596.8} & - & - & - & 17.00 & \multicolumn{1}{c}{3.52} & \multicolumn{1}{c|}{8608.4} & - & - & - \\
\multicolumn{1}{c|}{} & Patch-Rs\cite{Sparse-RS} & 32.25 & 6.56 & \multicolumn{1}{c|}{7257.0} & 6.00 & 12.69 & 47388.6 & 38.25 & 3.52 & \multicolumn{1}{c|}{6659.0} & 8.25 & 7.22 & 46345.8 \\
\multicolumn{1}{c|}{} & Ours & \textbf{100} & \textbf{6.56} & \multicolumn{1}{c|}{\textbf{1055.2}} & \textbf{100} & \textbf{12.38} & \textbf{1035.0} & \textbf{100} & \textbf{3.21} & \multicolumn{1}{c|}{\textbf{1129.3}} & \textbf{100} & \textbf{7.17} & \textbf{1058.3} \\ \bottomrule
\end{tabular}
}
\end{table*}

\subsection{Attacks on Face Verification}
In this section, we compare the attack performance of various black-box patch attacks on face verification. We set the query budgets for dodging and impersonation attacks to 10,000 and 50,000, respectively. The hyperparameters of DevoPatch are: $p=10$, $\mu=0.35$, $\gamma=1$. 
For dodging attacks, for randomly selecting a pair of faces with the same identity, the adversary generates an adversarial face to make the model recognize them as different identities. For impersonation attacks, the adversary generates an adversarial face to make the model recognize them as the same identity, which originally belonged to different identities. Here, we use cosine similarity and threshold to determine whether it is the same identity. When the cosine similarity of a pair of faces is greater than the threshold, the faces belong to the same identity.

Table~\ref{tab:face} shows decision-based patch attacks on face verification. Note that TPA exploits ILSVRC2012 on image classification to implement the attack through a class texture dictionary generated by style transfer. Here, because the face can directly represent the identity, we choose targeted images as the texture dictionary of TPA. These targeted images are the same as DevoPatch. AdvW and Patch-Rs achieve very few ASR in the dodging attack. Although HPA and MPA have extremely high ASR in the dodging attack, they tend to cover the face with a larger area and complete the attack with larger APA and extremely low query efficiency. Further, HPA, MPA, and Patch-Rs have very little ASR in the more challenging impersonation attack due to the limited information of the output of the label by the model. Because TPA has the targeted image as a prior, it achieves a good trade-off in ASR and ANQ in the dodging and impersonation attack. But even so, the performance of TPA in the impersonation attack can not achieve an extremely high ASR. Our DevoPatch and TPA share the same prior information (targeted images) in face verification, but DevoPatch significantly outperforms TPA in attack performance and query efficiency. In a dodging attack, the ANQ of TPA is usually three times that of DevoPatch. In the harder impersonation attack, the ANQ of DevoPatch is about one-twentieth of TPA. More importantly, in such a limited number of queries, DevoPatch has a smaller patch area and ASR, which is enough to illustrate the effectiveness of the proposed differential evolution patch attack algorithm.

\begin{figure*}[t]
\begin{minipage}[b]{1\linewidth}
  \centering
  \centerline{\includegraphics[width=18cm]{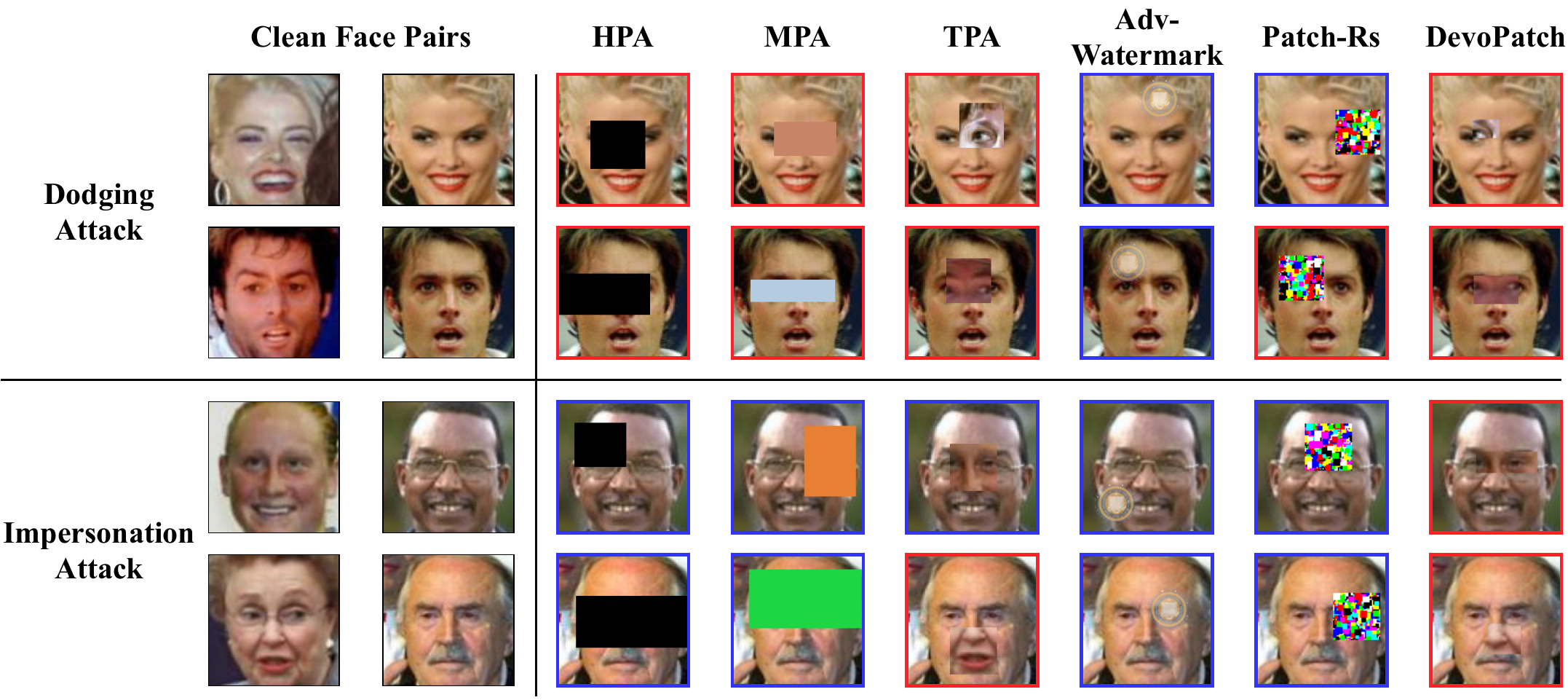}}
\end{minipage}
\caption{Visualization of patch attacks with ArcFace on face verification. The color of the face frame represents whether the attack is successful. Blue represents a failed attack, and red represents a successful attack. DevoPatch successfully achieves the attack of all examples with a small patch area.}
\label{fig:face_vis}
\end{figure*}

Table~\ref{fig:face_vis} shows the visualization of different patch attacks on face verification. Here, we choose ArcFace as the base model and visualize it on the LFW dataset. The color of the face frame represents whether the attack is successful. Blue represents a failed attack and red represents a successful attack. Because of the different semantic categories in image classification, the generated patches are easy to perceive. In face verification, since the color of patches is irrelevant to semantics, a similar situation also occurs in HPA, MPA, AdvW, and Patch-Rs. However, TPA and DevoPatch select faces as a prior and have face-related features, the resulting patches are relatively imperceptible. Further, since DevoPatch can better determine the location and shape of patches, it can improve attack performance and imperceptibility. DevoPatch has strong applicability and achieves query-efficient decision-based patch attacks on both image classification and face verification.

\begin{table}[t]
\centering
\caption{Attacks on the empirical and certifiable patch defenses.}
\label{tab:defense}
\scalebox{0.78}{
\begin{tabular}{@{}c|c|c|ccc|ccc@{}}
\toprule
\multirow{2}{*}{Type} & \multirow{2}{*}{Model} & \multirow{2}{*}{Method} & \multicolumn{3}{c|}{Untargeted Attack} & \multicolumn{3}{c}{Targeted Attack} \\ \cmidrule(l){4-9} 
 &  &  & ASR & APA & ANQ & ASR & APA & ANQ \\ \midrule
\multirow{6}{*}{Empirical} & \multirow{6}{*}{\makecell[c]{ResNet\\-152}} & Clean & 0 & - & - & 0 & - & - \\
 &  & Only Attack & 100 & 10.00 & 1349.7 & 100 & 23.78 & 1261.6 \\ \cmidrule(l){3-9} 
 &  & Only DW & 0.4 & - & - & 0.4 & - & - \\
 &  & Attack DW & 100 & 10.00 & 1346.8 & 100 & 23.78 & 1262.5 \\ \cmidrule(l){3-9} 
 &  & Only LGS & 3.1 & - & - & 3.1 & - & - \\
 &  & Attack LGS & 100 & 9.87 & 1383.8 & 100 & 23.63 & 1280.9 \\ \midrule
\multirow{4}{*}{Certifiable} & \multirow{2}{*}{\makecell[c]{ResNet\\-50}} & Only DS & 10.0 & - & - & 10.0 & - & - \\
 &  & Attack DS & 100 & 9.80 & 1159.1 & 100 & 22.72 & 1232.0 \\ \cmidrule(l){2-9} 
 & \multirow{2}{*}{ECViT-B} & Only ECViT & 2.9 & - & - & 2.9 & - & - \\
 &  & Attack ECViT & 100 & 17.43 & 1181.0 & 100 & 26.11 & 1138.7 \\ \bottomrule
\end{tabular}
}
\vspace{-0.1cm}
\end{table}

\subsection{Attacks on Patch Defenses}
We also evaluate the performance of DevoPatch against the patch defense methods on image classification, including Local Gradient Smoothing (LGS)\cite{LGS}, Digital Watermarking (DW)\cite{Dw}, Derandomized Smoothing (DS)\cite{DS} and  Efficient Certifiable Vision Transformer (ECViT)\cite{ECViT}. For empirical defenses, DW and LGS are regarded as pre-processing operations to remove adversarial patches. For certifiable defenses, we attack models including certifiable mechanisms. The backbone of DS is ResNet-50 and the backbone of ECViT is ECViT-B. Table~\ref{tab:defense} shows the adversarial robustness against empirical and certifiable patch defenses on DevoPatch. 
The above defenses cause very few images to be misclassified. For empirical patch defenses, DW and LGS do not take effect in the face of DevoPatch. A possible reason is that the adversarial patches produced by DevoPatch are part of natural images rather than adversarial perturbations generated by gradients. The former has semantics and harmony in visual understanding. For certifiable patch defenses, ECViT is the state-of-the-art certifiable patch defense, but it also cannot maintain certification in large rectangular patch areas (greater than 10\%). However, ECViT increases APA and reduces the quality of patches compared to DS. This experiment exposes deficiencies in existing patch defenses, so it is critical to improve the robustness and certification of defenses.

\begin{figure}[t]
\begin{minipage}[b]{0.49\linewidth}
  \centering
  \centerline{\includegraphics[width=4.5cm]{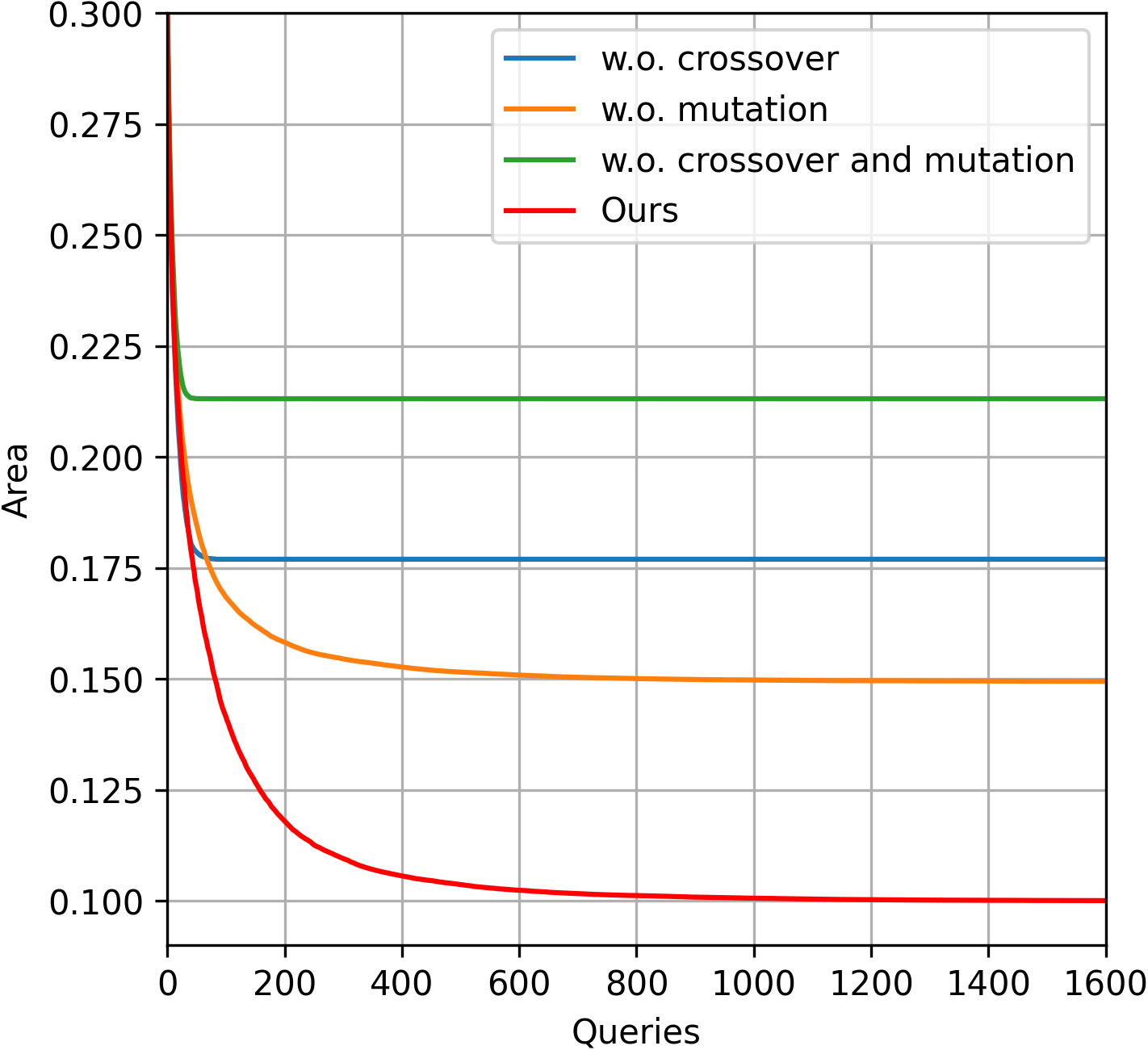}}
\centerline{\footnotesize (a) untargeted attack}
\end{minipage}
\hfill
\begin{minipage}[b]{0.49\linewidth}
  \centering
  \centerline{\includegraphics[width=4.5cm]{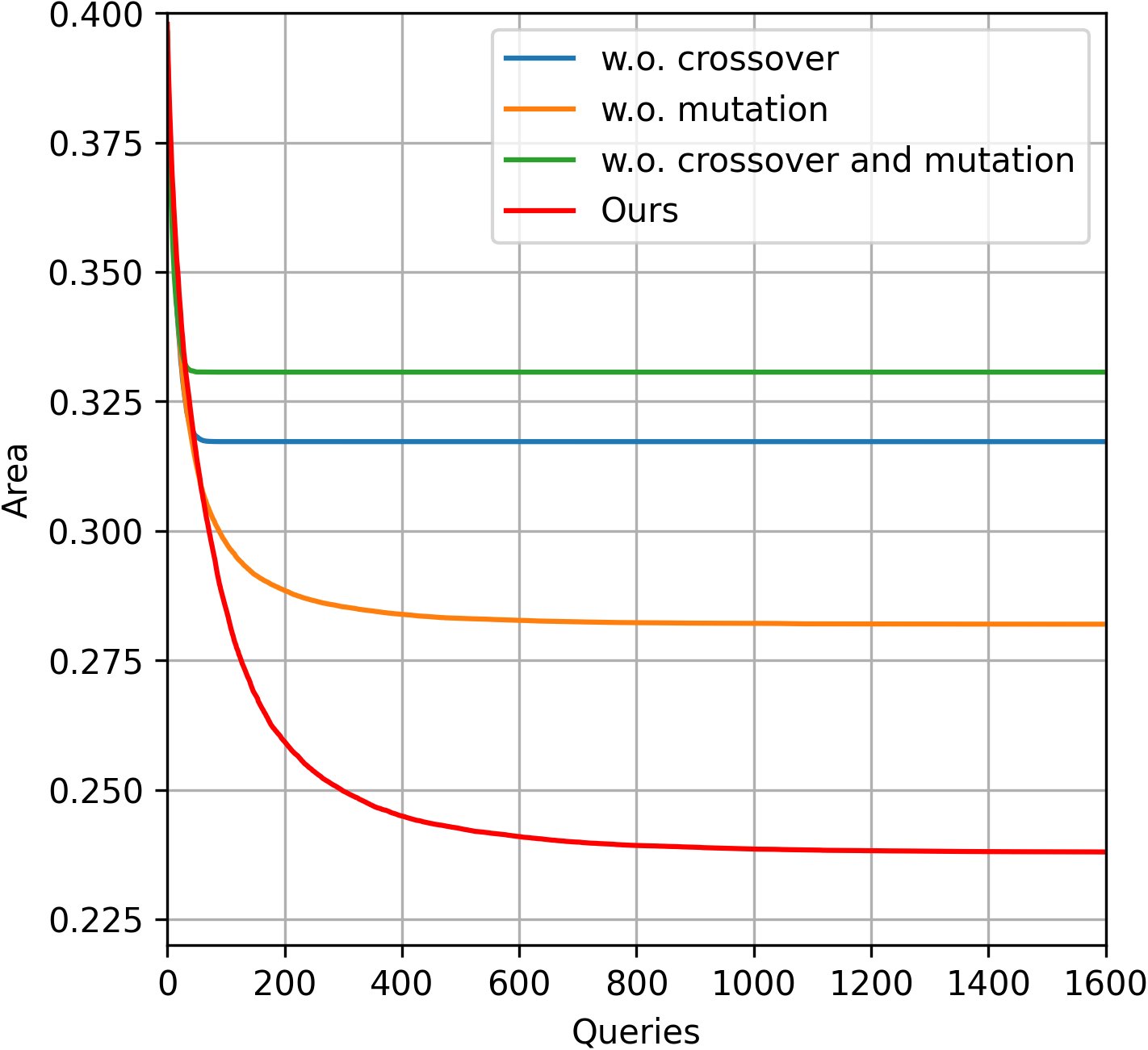}}
  \centerline{\footnotesize (b) targeted attack}
\end{minipage}
\caption{Ablation study on differential evolution. DevoPatch can better jump out of the local optimal solution and generate higher-quality adversarial patches.}
\label{fig:ablation_key}
\vspace{-0.1cm}
\end{figure}

\subsection{Ablation Study on Differential Evolution}
Both genetic algorithm (GA)~\cite{GA} and differential evolution algorithm (DE)~\cite{DE} are evolutionary algorithms, which simulate mutation, crossover, and selection in genetics to solve optimization problems. Due to different encoding, crossover, mutation, and selection strategies, DE generally has faster convergence speed~\cite{DE}. However, directly applying DE to this task encounters the challenges of complex solution space and efficient query efficiency. Therefore, we simplify solution spaces to the integer domain and improve the traditional DE.

We conduct ablation studies on differential evolution with ResNet-152, and the parameter settings are consistent with Section~\ref{sec:attack}. Fig.~\ref{fig:ablation_key} shows how APA changes as the number of queries increases under different differential evolutions. Here, \textit{w.o. crossover} and \textit{w.o. mutation} mean that the crossover and mutation improved by DevoPatch are not used, but the crossover and mutation of traditional DE~\cite{DE} are used. Under the premise of guaranteeing 100\% ASR, the mutation and crossover of DevoPatch have a fast convergence speed, and can better jump out of the local optimal solution, thereby generating higher-quality adversarial patches.

\begin{table}[t]
\centering
\caption{Analysis on different colors of target images.}
\label{tab:color}
\setlength{\tabcolsep}{3.0mm}
\begin{tabular}{@{}c|ccc|ccc@{}}
\toprule
\multirow{2}{*}{Color} & \multicolumn{3}{c|}{Untargeted Attack} & \multicolumn{3}{c}{Targeted Attack} \\ \cmidrule(l){2-7} 
 & ASR & APA & ANQ & ASR & APA & ANQ \\ \midrule
White & 100.0 & 15.26 & 1403.8 & 0.07 & 0.24 & 119.6 \\
Blue & 100.0 & 13.98 & 1333.5 & 0.10 & 0.33 & 122.6 \\
Green & 100.0 & 13.68 & 1296.6 & 0.12 & 0.44 & 123.2 \\
Yellow & 99.9 & 13.54 & 1351.3 & 0.07 & 0.25 & 120.6 \\
Pink & 100.0 & 15.06 & 1380.4 & 0.09 & 0.30 & 117.9 \\ \midrule
Ours & 100.0 & 10.00 & 1349.7 & 100.0 & 23.78 & 1261.6 \\ \bottomrule
\end{tabular}
\end{table}

\begin{table}[t]
\centering
\caption{Analysis on the randomness of target images.}
\label{tab:randomness}
\scalebox{0.85}{
\begin{tabular}{@{}c|ccc|ccc@{}}
\toprule
\multirow{2}{*}{Random} & \multicolumn{3}{c|}{Untargeted Attack} & \multicolumn{3}{c}{Targeted Attack} \\ \cmidrule(l){2-7} 
 & ASR & APA & ANQ & ASR & APA & ANQ \\ \midrule
(1) & 100.0 & 10.00 & 1349.7 & 100.0 & 23.78 & 1261.6 \\
(2) & 100.0 & 9.85 & 1348.7 & 100.0 & 22.78 & 1238.0 \\
(3) & 100.0 & 9.86 & 1357.4 & 100.0 & \textbf{22.19} & 1227.8 \\
(4) & 100.0 & \textbf{9.81} & 1350.2 & 100.0 & 22.52 & 1223.1 \\
(5) & 100.0 & 9.83 & \textbf{1331.9} & 100.0 & 22.57 & \textbf{1220.6} \\ \midrule
Mean & \textbf{100.0} & \textbf{9.86±0.07} & \textbf{1347.6±9.4} & \textbf{100.0} & \textbf{22.77±0.60} & \textbf{1234.2±16.6} \\ \bottomrule
\end{tabular}
}
\vspace{-0.1cm}
\end{table}

\begin{table}[t]
\centering
\caption{Analysis of different sources of target images.}
\label{tab:source}
\scalebox{0.9}{
\begin{tabular}{@{}c|c|ccc|ccc@{}}
\toprule
\multirow{2}{*}{Model} & \multirow{2}{*}{Data Sources} & \multicolumn{3}{c|}{Untargeted Attack} & \multicolumn{3}{c}{Targeted Attack} \\ \cmidrule(l){3-8} 
 &  & ASR & APA & ANQ & ASR & APA & ANQ \\ \midrule
\multirow{2}{*}{ResNet} & Same & 100 & 10.77 & 1262.9 & 100 & 24.10 & 1256.9 \\
 & Different & 100 & 10.49 & 1374.2 & 100 & 25.55 & 1210.1 \\ \midrule
\multirow{2}{*}{ViT-B} & Same & 100 & 15.79 & 1359.3 & 100 & 29.80 & 1257.3 \\
 & Different & 100 & 14.83 & 1332.2 & 100 & 26.34 & 1236.6 \\ \midrule
\multirow{2}{*}{Mixer-B} & Same & 100 & 15.01 & 1240.9 & 100 & 25.73 & 1229.6 \\
 & Different & 100 & 15.91 & 1302.4 & 100 & 25.43 & 1298.6 \\ \bottomrule
\end{tabular}
}
\end{table}

\subsection{Analysis on Target Images} 
\label{sec:targetimage}
To explore the impact of the selection of target images on attack performance, we conduct experimental analysis on image classification with ResNet-152 from three perspectives, including color, randomness, and data source. The parameter settings are consistent with Section~\ref{sec:attack}.

\textbf{Color.} HPA~\cite{HPA} and MPA~\cite{patchattack} introduce monochrome patches to implement the attack. Therefore, monochrome images have the potential to become target images. Here, we select \textit{White}, \textit{Blue}, \textit{Green}, \textit{Yellow} and \textit{Pink} as target images. Table~\ref{tab:color} illustrates the attack performance and query efficiency when images of different colors are used as target images. Under the untargeted setting, DevoPatch with monochrome images has a similar ASR and APA as MPA, but the query efficiency is one-seventh of MPA. However, monochrome images have almost no target attack performance, because monochrome images have almost no semantic information of the corresponding class. The above experiment shows the efficiency of DevoPatch and the necessity of random natural images as target images.

\textbf{Randomness.} Considering that target images on image classification are randomly selected from the ILSVRC2012 validation set, randomness may affect attack performance and query efficiency. Here, we fix the clean images and randomly sample the target images five times, then evaluate the performance. Table~\ref{tab:randomness} illustrates the impact of different random target images on attack performance and query efficiency. From the experiments, we can find that although the marginal improvement can be obtained through randomness, the impact of random target images on performance is very small, and APA and ANQ are very close. Although the optimal performance is randomly selected multiple times, the query cost is multiplied, which is not feasible in real-world scenarios. From the perspective of the target images themselves, how to generate a more powerful target image is a future work that has the potential to improve the attack efficiency.

\label{sec:source}
\textbf{Data Source.} This work is carried out in a black-box decision-based setting. For the black-box model, we can only obtain the output label, which is difficult to obtain full training data or access the model architecture.
As described in Section~\ref{sec:DevoPatch}, we use targeted image priors to reduce the complexity of the solution space and achieve effective targeted attacks. In Section~\ref{sec:attack}, we choose the targeted images randomly sampled from the validation set of ILSVRC2012 and show the effectiveness of DevoPatch, which belong to the same source data as the training set of the models. 
To further demonstrate the generalization capability of the proposed method in real-world scenarios, we collect 100 images from the Internet as the targeted images, which are not from the same source as ILSVRC2012.
In the case of using different-source data, ASR equivalent to same-source data can be obtained on image classification, as described in Table~\ref{tab:source}. DevoPatch based on different-source data has very subtle differences in areas and queries, which shows DevoPatch is not sensitive to the domain of targeted images. 
It is worth noting that TPA uses ImageNet to generate a texture dictionary to attack the classification model, which is impossible in real-world scenarios. However, DevoPatch can arbitrarily select a correctly identified image from the Internet as the targeted image, thereby realizing a black-box patch attack with high operability and flexibility.

\begin{figure*}[t]
\begin{minipage}[b]{1\linewidth}
  \centering
  \centerline{\includegraphics[width=18cm]{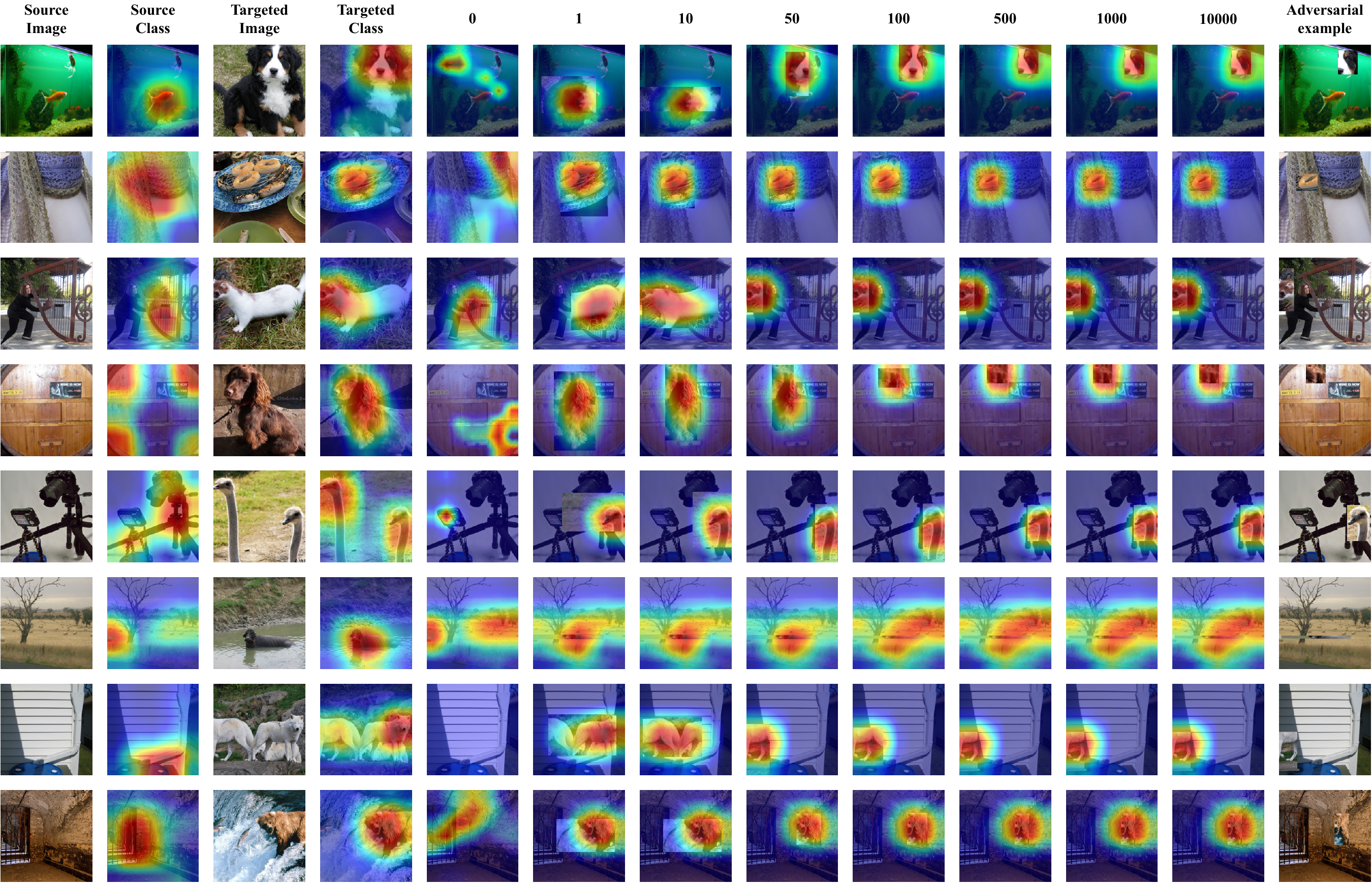}}
\end{minipage}
\caption{Demonstration of the different discriminative regions of ResNet-152 on image classification. We utilize the gradient-based class activation mapping\cite{gradcam} to visualize the attention maps of various classes. Among the limited queries, we notice that the class activation map of the class predicted by the model focuses on the adversarial patch, indicating that DevoPatch can become the most discriminative region without having access to any details of the model. Since adversarial patches based on targeted classes cover the most discriminative regions, the model outputs predictions for the targeted class.}
\label{fig:attention}
\end{figure*}

\subsection{Effectiveness Analysis}
In this section, we analyze why DevoPatch has a very high targeted attack success rate. We utilize the gradient-based class activation mapping (Grad-CAM)\cite{gradcam} to visualize the attention maps of various classes, as shown in Figure~\ref{fig:attention}. First, we use Grad-CAM to generate the attention maps of the source image and targeted image of their corresponding classes (such as column 2 and column 4). We can find that the most discriminative regions are all in the regions with the most salient category objects. Then, we visualize the attention map of the source image corresponding to the target class and find that it is not focused on the objects of the source class. Among the limited queries, we notice that the class activation map of the class predicted by the model focuses on the adversarial patch, indicating that DevoPatch can become the most discriminative region without having access to any details of the model. Since adversarial patches based on targeted classes cover the most discriminative regions, the model outputs predictions for the targeted class. Therefore, DevoPatch is a query-efficient decision-based patch attack because of paired key-points and targeted image prior.

\section{Discussion}
\label{sec:discussion}
In this section, we compare the robustness of ResNet, ViT, and MLP models to patch perturbation on image classification. In Table~\ref{tab:black}, we find that our method needs a relatively larger area to craft successful patch attacks on ViT and MLP models compared with ResNet model with similar query budgets. It means ViT and MLP models are relatively more robust than ResNet model under the most threatening decision-based patch attack. 
Probably because ViT and MLP split the image into multiple non-overlapping patches which reduce the impact of noises on one local region to the final classification results\cite{DBLP:journals/corr/abs-2103-14586}. 
As shown in Table~\ref{tab:transfer}, all kinds of models are equally vulnerable to perturbations computed using white-box attack GAP. We then find that adversarial perturbations computed using ResNet rarely transfer to ViT or MLP in the white-box setting especially for the targeted attack, which is also observed by \cite{DBLP:journals/corr/abs-2103-14586} in imperceptible attacks. Interestingly, different from the conclusion in \cite{DBLP:journals/corr/abs-2103-14586}, we find that adversarial perturbations computed using ViT and MLP do transfer to ResNet. 
Particularly, the adversarial perturbations crafted by our black-box method transfer more easily over different architectures, even for the targeted attack.
In addition, we first present the lower bound on the area required for the targeted patch attack in the decision-based setting. Targeted attacks of any category can be completed in about 25\% of the patch area under about 1,300 queries. This is an important safety reference for real-world systems.
The above observations suggest that studying the adversarial robustness of DNNs from the perspective of decision-based black-box patch attacks is necessary to better understand and improve DNNs.

\section{Conclusions}
\label{sec:conclusions}
In this work, we explore the practical threat of decision-based black-box patch attack to the robustness of existing DNNs for the first time. Compared with transfer-based and score-based settings, decision-based settings do not require access to a large amount of training data and only rely on minimal information, the labels by the model’s output, to achieve the adversarial attack. To simplify the solution space and improve query efficiency, we propose a differential evolutionary algorithm named DevoPatch for query-efficient adversarial patch attacks in the decision-based black-box setting. In DevoPatch, we model adversarial patches as paired key-points and utilize targeted images as priors. With paired key-points and targeted image priors, the differential evolution algorithm based on the integer domain greatly improves the query efficiency.
As a result of our comprehensive results, DevoPatch outperforms the state-of-the-art black-box patch attack in terms of patch area and ASR both on image classification and face verification. 
More importantly, with a reduced solution space, DevoPatch illustrates significant query-efficiency when compared with the existing patch attacks in the decision-based black-box setting. 
We also investigate the robustness of various DNN architectures against DevoPatch.

DevoPatch exposes the shortcomings of existing DNNs against patch attacks. In future research, we can use DevoPatch to evaluate the robustness of the model to black-box adversarial patches. In addition, our work provides a deep understanding of the robustness of DNNs against decision-based patch attacks. We believe this work could be used to inform and inspire the design and deployment of robust vision models based on various DNN architectures in the future.

\section*{Acknowledgments}
This work was supported by National Natural Science Foundation of China (No.62072112), Scientific and Technological innovation action plan of  Shanghai Science and Technology Committee (No.22511102202), Fudan Double First-class Construction Fund (No. XM03211178).

\bibliography{citation}
\bibliographystyle{IEEEtran}
\end{document}